\documentclass[lettersize,journal]{IEEEtran}
\usepackage{graphicx} 
\usepackage{amsmath}
\usepackage{amssymb}
\usepackage{booktabs}
\usepackage{color}
\usepackage{graphicx} 
\usepackage{pdfpages} 

\usepackage{enumerate}

\usepackage{amssymb}
\usepackage{amsmath}
\usepackage{subfigure}
\usepackage{setspace}
\usepackage{float}
\newcommand{\modelname}{RS-MoE}

\newcommand{\revise}[1]{\textcolor{black}{#1}}

\newcommand{\minorrevise}[1]{\textcolor{black}{#1}}

\title{RS-MoE: A Vision-Language Model with Mixture of Experts for Remote Sensing Image Captioning and Visual Question Answering}

\author{Hui Lin\textsuperscript{\textdagger}, Danfeng Hong\textsuperscript{\textdagger}, \IEEEmembership{Senior Member, IEEE}, Shuhang Ge\textsuperscript{\textdagger}, \\ Chuyao Luo, Kai Jiang, Hao Jin, and Congcong Wen* \IEEEmembership{Member, IEEE}
\thanks{This work was supported in part by the National Key Research and Development Program of China (2021YFC3300500) and Beijing Nova Program (2024124). (Hui Lin, Danfeng Hong, and Shuhang Ge contributed equally to this work). \textit{(Corresponding author: Congcong Wen)}.}
\thanks{Hui Lin, Kai Jiang, and Hao Jin are with China Academy of Electronics and Information Technology, Beijing 100846, China. (e-mail: linhui@whu.edu.cn, kaijiang@mail.ustc.edu.cn, and jinhao1@cetc.com.cn) }
\thanks{D. Hong is with the Aerospace Information Research Institute, Chinese Academy of Sciences, 100094 Beijing, China, and also with the School of Electronic, Electrical and Communication Engineering, University of Chinese Academy of Sciences, Beijing 100049, China. (e-mail: hongdf@aircas.ac.cn).}

\thanks{Chuyao Luo is with the Department of Computer Science, Harbin Institute of Technology, Shenzhen 518055, China. (e-mail: luochuyao.dalian@gmail.com).}

\thanks{Shuhang Ge and Congcong Wen are with the Department of Electrical and Computer Engineering, New York University Abu Dhabi, Abu Dhabi, UAEe-mail: sg7484@nyu.edu and wencc@nyu.edu).}
}

\begin{document}

\maketitle

\begin{abstract}
Remote Sensing Image Captioning (RSIC) presents unique challenges and plays a critical role in applications such as environmental monitoring, urban planning, and disaster management. Traditional RSIC methods often struggle to produce rich and diverse descriptions. Recently, with significant advancements in Vision-Language Models (VLMs), efforts have emerged to integrate these models into the remote sensing domain and to introduce richly descriptive datasets specifically designed to enhance VLM training. However, most current RSIC models generally apply only fine-tuning to these datasets without developing models tailored to the unique characteristics of remote sensing imagery. This paper proposes RS-MoE, the first Mixture of Expert based VLM specifically customized for remote sensing domain. Unlike traditional MoE models, the core of RS-MoE is the MoE Block, which incorporates a novel Instruction Router and multiple lightweight Large Language Models (LLMs) as expert models. The Instruction Router is designed to generate specific prompts tailored for each corresponding LLM, guiding them to focus on distinct aspects of the RSIC task. This design not only allows each expert LLM to concentrate on a specific subset of the task, thereby enhancing the specificity and accuracy of the generated captions, but also improves the scalability of the model by facilitating parallel processing of sub-tasks. Additionally, we present a two-stage training strategy for tuning our RS-MoE model to prevent performance degradation due to sparsity. We fine-tuned our model on the RSICap dataset using our proposed training strategy. Experimental results on the RSICap dataset, along with evaluations on other traditional datasets where no additional fine-tuning was applied, demonstrate that our model achieves state-of-the-art performance in generating precise and contextually relevant captions. Notably, our RS-MoE-1B variant achieves performance comparable to 13B VLMs, demonstrating the efficiency of our model design. Moreover, our model demonstrates promising generalization capabilities by consistently achieving state-of-the-art performance on the Remote Sensing Visual Question Answering (RSVQA) task.
\end{abstract}


\section{Introduction}

Image captioning integrates computer vision and natural language processing to automatically generate descriptive text for images, effectively bridging the information gap between visual and linguistic domains. Unlike captioning of common natural images, remote sensing image captioning (RSIC) poses greater challenges due to the inherent characteristics of remote sensing images, which cover larger geographic areas and more diverse geographical objects~\cite{lin2025generalization,chen2020road}. Consequently, RSIC tasks demand not only the description of geographic object information but also a comprehensive understanding of relationships and interactions among various geographic objects.

\revise{Early studies of RSIC focused on traditional methodologies, including template-based ~\cite{shi2017can} and retrieval-based methods~\cite{gong2014improving,sun2015automatic}.
However, these methods cannot generate rich and varied descriptive sentences. Recent studies~\cite{shen2020remote, zhao2021high, zia2022transforming,ye2022joint,yang2022meta,wei2023vlca} commonly adopt the encoder-decoder architecture, which divides the RSIC task into an image encoding phase that extracts semantic features from the input image, and a sequence modeling phase that uses the extracted features to generate text and sentences. Depending on the specific models employed, encoder-decoder approaches can be further categorized into CNN-based encoders with RNN/LSTM decoders~\cite{zhao2021high}, and CNN-based encoders with Transformer decoders~\cite{zia2022transforming}. Although these methods have achieved satisfying performance on the RSIC task, they typically generate only one or two simple sentences for captioning, limiting their practical application. This limitation typically arises due to two principal reasons: the simplicity and repetitiveness of the sentences in the datasets used for training models; and the relatively limited ability of these models to extract semantic features and generate complex descriptions.}

The Large Language Models (LLMs) and Vision Language Models (VLMs) have recently achieved significant success across multiple fields, including computer vision~\cite{radford2021learning}, natural language processing~\cite{zhou2022learning}, and robotics~\cite{wen2024secure,wen2025zero}. Particularly, VLMs~\cite{li2023blip, zhu2023minigpt,dai2023instructblip} have effectively narrowed the gap between visual images and textual natural language by advancing the understanding of intermodal relationships, reaching a level of visual comprehension comparable to human capabilities. Therefore, some researchers~\cite{li2023rs,hu2023rsgpt,kuckreja2023geochat,pang2024h2rsvlm,zhang2024earthgpt,hong2024spectralgpt,hong2024multimodal,li2024seamo} have recently shifted their focus to applying VLMs to remote sensing visual interpretation tasks. Specifically, for RSIC tasks, the RSGPT~\cite{hu2023rsgpt} introduced the human-annotated RSICap dataset, which provides high-quality and detailed descriptions for remote sensing images. The remote sensing captions in this dataset include summarized scene descriptions, object information (color, shape, and count), and object relationships (relative position). However, RSGPT merely fine-tunes an existing VLM model on this dataset without proposing a novel model to explore and utilize the provided detailed descriptions deeply.

{In this paper, we propose a novel Vision-Language Model (VLM) based on the Mixture of Experts (MoE) framework, named RS-MoE, specifically customized for the Remote Sensing Image Captioning (RSIC) task. To the best of our knowledge, this is the first work that applies the MoE framework to VLMs in the remote sensing domain. We extend the core principles of MoE to specifically address the unique challenges of remote sensing images. By decomposing the RSIC task into specialized subtasks, we leverage different expert models to facilitate more precise and contextually aware caption generation. Specifically, RS-MoE consists of three key components: the Image Encoder, the VLM Encoder, and the MoE Block. Unlike conventional MoE modules, our MoE Block consists of \revise{\textit{a novel Instruction Router}} and \revise{\textit{multiple lightweight LLMs}}. The Instruction Router is designed to generate tailored prompts for each LLM. In our experiments, these prompts focus on theme comprehension, object recognition, and relationship inference. The Instruction Router dynamically adjusts prompts based on visual features and task instructions, offering a more precise level of control and enabling the model to generate descriptions specifically tailored to each captioning objective. Furthermore, unlike traditional MoE models that typically use feed-forward networks (FFNs) as experts, we incorporate lightweight LLMs as expert models. This design allows each LLM to process complex, task-specific linguistic information efficiently and generate detailed, contextually aware descriptions while ensuring computational feasibility.  In addition to the RS-MoE model, we propose a \revise{\textit{two-stage training strategy}} to fine-tune the MoE framework for the RSIC task, addressing potential model degradation caused by the sparsity that typically arises when applying MoE to remote sensing images. This sparsity arises from the lack of representation of remote sensing-specific features in models pretrained on natural images, leading to inefficiencies in capturing the complex characteristics of remote sensing imagery, which ultimately affects model performance. Moreover, to significantly reduce the number of trainable parameters, we incorporate the Low-Rank Adaptation (LoRA) training strategy into the RS-MoE tuning stages. After fine-tuning our model on the RSICap dataset using our proposed training strategy, we evaluated it on the RSICap dataset and directly tested it on traditional datasets, such as UCM-Captions, Sydney-Captions, and RSICD, without any additional fine-tuning on these datasets. Experimental results demonstrate that our RS-MoE-7B model achieves state-of-the-art performance, while the RS-MoE-1B variant achieves performance comparable to 13B VLMs in generating precise and contextually relevant captions. Furthermore, we extend our model to the Remote Sensing Visual Question Answering (RSVQA) task and demonstrate promising generalization capabilities by consistently achieving state-of-the-art performance on this task. Our contributions are summarized as follows:}

\begin{itemize}



\item{\revise{We are the first to introduce the MoE framework to multimodal remote sensing, leveraging its task decomposition concept with specialized expert models to effectively address the complexity and diversity of remote sensing data across both visual and textual modalities.}}


\item{ \revise{We propose RS-MoE, a novel Vision-Language Model specifically designed for remote sensing image captioning, with an Instruction Router that dynamically generates task-specific prompts and lightweight LLMs as expert models to enhance both effectiveness and efficiency.}}


\item{ \revise{We present a two-stage training strategy for RS-MoE, incorporating proper initialization to mitigate sparsity-induced degradation and employing LoRA to reduce trainable parameters, improving both efficiency and manageability of the model during the training process.}}


\item{ \revise{Extensive experiments demonstrate that, with fine-tuning on a single dataset $\sim 3,000$ images, our model achieves state-of-the-art performance on five RSIC datasets and strong generalization on two RSVQA datasets. Notably, the lightweight RS-MoE-1B matches the performance of larger 13B VLMs while being significantly more efficient.}}

\end{itemize}
\begin{figure*}[h]
    \centering
    \includegraphics[width=\linewidth]{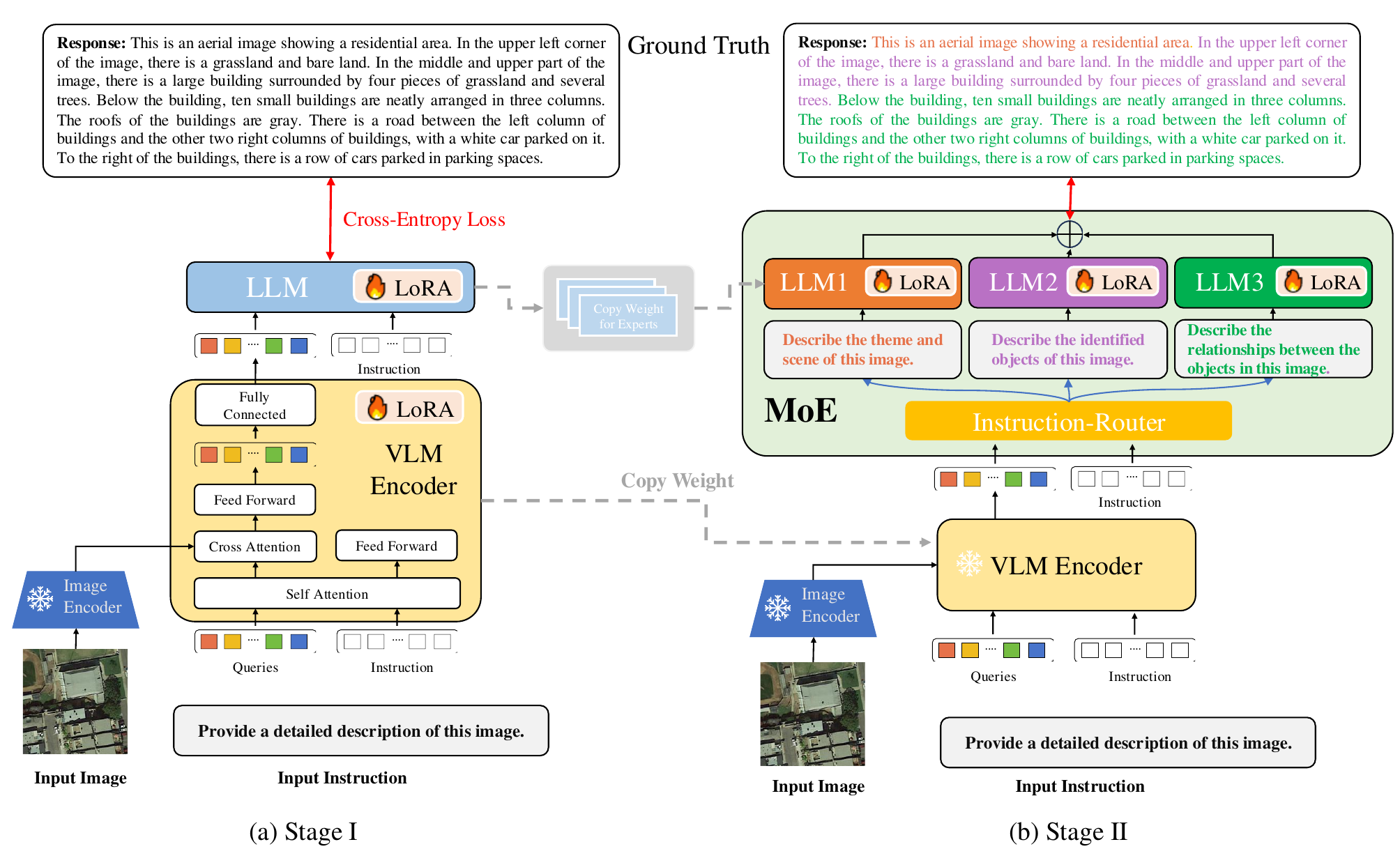}
    \caption{\revise{Overview of the proposed RS-MoE model, which consists of four key components: the Image Encoder, the VLM Encoder, the LLM Block and the MoE Block. The MoE Block comprises an Instruction Router that dynamically generates task-specific prompts and three lightweight LLMs, which focus on different aspects of the captioning task. In the generated captions, shown in the top right corner of the figure, distinct colors represent each aspect: orange for the overall theme, purple for specific objects, and green for relationships between objects. RS-MoE is trained using a novel two-stage training strategy specifically designed for remote sensing image captioning. In Stage I (a), the VLM Encoder and the LLM Block are fine-tuned to initialize model weights specifically designed for the RSIC task. In Stage II (b), the MoE Block is fine-tuned to produce more detailed captions for RSIC tasks.}}
    \label{fig_moe}
\end{figure*}

\section{Related Work}

\subsection{Remote Sensing Image Captioning}
Remote Sensing Image Captioning (RSIC) poses a challenging task within the fields of computer vision and specialized linguistic modeling, aiming to generate descriptive narratives for remote sensing images by focusing on their visual attributes. Early methodologies predominantly employed template-based ~\cite{shi2017can} and retrieval-based techniques ~\cite{gong2014improving,sun2015automatic}. Template-based strategies involve creating fixed sentence structures and populating them with relevant information, such as attributes and categories displayed within the image. Conversely, retrieval-based strategies develop large-scale databases containing images and their corresponding textual descriptions. These strategies retrieve the most similar image from the database and provide the corresponding description.

Recent studies~\cite{shen2020remote, zhao2021high, zia2022transforming, wei2023vlca} have proposed methods based on the Encoder-Decoder framework for achieving RSIC tasks. This architecture typically incorporates an image encoder to extract semantic features from images, which are then interpreted by a sequential model to produce textual annotations. Depending on the model used for the encoder, existing Encoder-Decoder methodologies for RSIC can be categorized into two types: CNN-based image encoders with RNN-based text decoders, and CNN-based image encoders with Transformer-based text decoders. For instance, ~\cite{shen2020remote} first fine-tuned the CNN alongside the VAE in the first phase, followed by employing both Transformer and Reinforcement Learning to process spatial and semantic features for text generation. To incorporate the domain-specific knowledge of remote sensing, ~\cite{zhao2021high} introduced a fine-grained and structured attention-based model that capitalizes on the structural attributes of semantic content in high-resolution RS images. Besides, ~\cite{zia2022transforming} first implemented a multi-scale visual feature encoder to extract detailed information from RS images and then engaged an adaptive multi-head attention decoder to refine the text generation process based on the extracted multi-scale features.

More recently, some researchers have begun to explore the use of VLM for the RSIC task. ~\cite{wei2023vlca} introduced a vision-language aligning network that utilizes CLIP for visual feature extraction and a pretrained generative language model, GPT-2, to generate relevant descriptions for given remote sensing images. Additionally, ~\cite{hu2023rsgpt} have developed a high-quality, human-annotated remote sensing image captioning dataset with detailed and rich descriptions specifically for the RSIC task, as well as an evaluation dataset for assessing general VLMs. They further fine-tuned the Instruct-BLIP model on this dataset, establishing a benchmark for both RSIC and RSVQA tasks.

{\color{black}
\subsection{VLM in Remote Sensing}

Vision-Language Models (VLMs)~\cite{li2023blip, zhu2023minigpt, dai2023instructblip} are advanced frameworks that integrate visual and textual modalities, enabling the interpretation and generation of visual content with corresponding linguistic descriptions. These models have achieved remarkable progress in tasks involving natural images, particularly in generating dialogues and interpreting visual data. By combining visual features with contextual cues, VLMs offer an interactive and coherent understanding of visual inputs~\cite{liu2024visual, wang2024qwen2}.

Inspired by the rapid advancements in computer vision, researchers have increasingly focused on integrating Vision-Language Models (VLMs) into remote sensing tasks. One of the early comprehensive reviews~\cite{wen2023vision}, highlighted key VLM applications within the remote sensing domain, identifying both the current challenges and future directions for research. Since then, several VLM-based models have emerged to address these challenges~\cite{li2023rs,hu2023rsgpt,kuckreja2023geochat,pang2024h2rsvlm,zhang2024earthgpt,hong2024spectralgpt,hong2024multimodal,lin2025fedrsclip}. For example, ~\cite{zhang2024earthgpt} introduced EarthGPT, a model specifically designed for multi-sensor remote sensing image interpretation. This model integrates three core innovations: vision-enhanced perception, cross-modal understanding, and a unified instruction tuning mechanism, making it particularly effective for handling complex remote sensing data. In addition to these novel approaches, many remote sensing VLMs build upon the foundational architecture of LLaVA, adapting it to better address the specific challenges of remote sensing. RS-LLaVA enhances the standard LLaVA framework by incorporating a low-rank adaptation (LoRA) mechanism, optimizing the model's ability to process the multi-scale and high-resolution nature of remote sensing images while maintaining computational efficiency. Similarly, H2RSVLM~\cite{pang2024h2rsvlm} extends LLaVA by integrating a pretrained vision encoder and LLM, connected through a multilayer perceptron (MLP) projector. This model is particularly designed to improve its handling of complex queries and mitigate incorrect generations, especially in remote sensing contexts where "hallucinations" are a common issue. GeoChat~\cite{kuckreja2023geochat}, also based on LLaVA, incorporates task-specific prompts and spatial capabilities, enabling it to handle multiple remote sensing tasks more effectively. These enhancements allow the model to better understand visual prompts and generate grounded object descriptions, providing more detailed and context-specific interactions in response to image and region-level queries.}

{\color{black}
\subsection{Mixture of Experts}
The Mixture of Experts (MoE) framework, first introduced by~\cite{moe}, is a foundational technique in machine learning designed to dynamically allocate computational resources by selectively activating specialized experts for specific tasks. This approach has been extensively studied across diverse domains, including recommender systems, natural language processing, computer vision, and multimodal learning. For example, in recommender systems, the Multi-gate Mixture of Experts (MMoE)~\cite{mmoe} was developed as a multi-task learning model that explicitly models task relationships by sharing expert submodels across tasks, with separate gating networks optimizing individual tasks. Similarly, in natural language processing, the Stable Transferable Mixture of Experts (ST-MoE)~\cite{zoph2022st} enhances the practicality and robustness of sparse models, achieving state-of-the-art performance on a wide range of natural language benchmarks. In computer vision, the Vision Mixture of Experts (V-MoE)~\cite{riquelme2021scaling} introduced a sparsely activated transformer model capable of scaling to 15 billion parameters, setting a new benchmark for efficiency and scalability in vision tasks. In the multimodal domain, the Language-Image Mixture of Experts (LIMoE)~\cite{mustafa2022multimodal} integrates vision and language modalities through contrastive training, enabling expert layers to dynamically partition data based on modalities. Furthermore, the MoE-LLaVA~\cite{lin2024moe} activates only the top-k experts during inference, optimizing computational efficiency for large vision-language models.

In contrast, the application of MoE in remote sensing remains largely underexplored. For instance, the MV-MoE~\cite{cao2024mv} leverages a multi-task learning framework by incorporating attention mechanisms to represent multi-modal images across various scenes adaptively. Similarly, the SW-MoE~\cite{he2024mixture} employs an MoE structure to activate experts based on different image scales and scenes, addressing challenges such as complex backgrounds and imbalanced datasets. Despite these advancements, existing studies are confined to single-modality applications. To the best of our knowledge, no prior work has explored the use of MoE in remote sensing for multimodal tasks. In this work, we address this gap by proposing a novel MoE-based VLM specifically designed to tackle the unique challenges of remote sensing image captioning and VQA.

}



\section{Methods}

\subsection{Problem Statement}

Let \( I \in \mathbb{R}^{W \times H \times 3} \) be a remote sensing image, where \( W \) and \( H \) denote the width and height of the image, respectively, and the \( 3 \) represents the RGB color channels. The objective of RSIC is to generate a descriptive caption \( S = \{s_1, \cdots, s_n\} \), where each \( s_i \) is a sentence that effectively summarizes the content of the image. Consider \( \mathcal{O} = \{o_1, o_2, \ldots, o_m\} \) as the set of visible geographic objects within \( I \), where each \( o_j \) represents an object identified in the remote sensing image. The relationships among these objects can be denoted as \( \mathcal{R} \), where \( \mathcal{R} \subseteq \mathcal{O} \times \mathcal{O} \) represents a set of tuples \((o_i, o_k)\), indicating a significant interaction or spatial relationship between object \( o_i \) and object \( o_k \). An effective captioning model for RSIC should not only \textit{identify the objects} in \( \mathcal{O} \) but also \textit{interpret the relationships} in \( \mathcal{R} \).

{\color{black}
\subsection{Model Architecture} \label{sec:model}
As shown in Fig.~\ref{fig_moe}, the proposed RS-MoE model consists of four modules, including an Image Encoder, a Vision Language Model (VLM) Encoder, a Large Language Model (LLM) Block, and a Mixture of Experts (MoE) Block.

\subsubsection{Image Encoder}

\revise{Given the remote sensing image \( I \) as defined earlier, we employ an Image Encoder \( f_I \) to extract its visual features \( F_I \). Specifically, \( F_I = f_I(I) \), where \( F_I \in \mathbb{R}^{C \times W' \times H'} \), \( C \) is the number of feature channels, and \( W' \) and \( H' \) represent the downsampled spatial dimensions. For this purpose, we adopt ViT-G/14~\cite{zhai2022scaling}, a Vision Transformer with nearly two billion parameters, as the backbone of the Image Encoder. To prevent overfitting and ensure stable feature extraction, the parameters of \( f_I \) are frozen during the entire training process.}

\subsubsection{VLM Encoder} Inspired by InstructBLIP~\cite{dai2023instructblip}, our VLM Encoder, denoted as \( f_{\text{VLM}} \), is specifically designed to generate visual features aligned with the given instructions \( T \). The process begins with a set of learnable query embeddings \( Q_T \in \mathbb{R}^{L \times C'} \), where \( L \) is the number of queries and \( C' \) is the feature dimension. These queries, combined with the instructions \( T \), are processed through the self-attention layers of the encoder to produce an intermediate instruction-aware feature set \( F_{\text{SA}} \), defined as:
\begin{equation}
    F_{\text{SA}} = \textit{SelfAttention}(Q_T, T)
\end{equation}
Next, the features \( F_{\text{SA}} \) are integrated with the visual features \( F_I \) extracted by the Image Encoder using cross-attention layers, resulting in instruction-aware visual features \( F_{\text{CA}} \):
\begin{equation}
    F_{\text{CA}} = \textit{CrossAttention}(F_{\text{SA}}, F_I)
\end{equation}
This step dynamically aligns the visual features with the task-specific instructions, tailoring them to the input context. Finally, the refined features \( F_{\text{CA}} \) pass through a feed-forward network and a fully connected layer to yield the final feature representation \( F_{\text{VLM}} \):
\begin{equation}
    F_{\text{VLM}} = \text{FFN}(F_{\text{CA}})
\end{equation}
By incorporating these steps, the VLM Encoder ensures that the extracted features are both adaptable and insightful, effectively aligning visual and instruction-specific information to support downstream tasks.

\subsubsection{LLM Block} In the first training stage, The learned instruction-aware features \( F_{\text{VLM}} \), along with the corresponding instructions \( T \), serve as input prompts to the LLM Block \( f_l \). The LLM Block generates a descriptive caption \( S = \{s_1, \cdots, s_n\} \) that effectively summarizes the content of the remote sensing image \( I \) by identifying the objects in \( \mathcal{O} \) and interpreting their relationships in \( \mathcal{R} \). Formally, this process can be expressed as:
\begin{equation}
    S = f_l(F_{\text{VLM}}, T)
\end{equation}
In the first training stage, the LLM Block is fine-tuned to initialize the weights of the LLM models used in the MoE Block. This preparatory step ensures that the models are capable of accurately capturing both object-level details and relationship dynamics, equipping the system to generate precise and coherent captions in subsequent stages.

\subsubsection{MoE Block} Similarly, in the second training stage, the features generated by the VLM Encoder, \( F_{\text{VLM}} \), are directly input into the MoE Block to facilitate caption generation. Unlike traditional MoE modules, the MoE Block comprises a novel instruction-router \( R \) and \( N \) distinct LLMs, denoted as \( \{f_{l1}, f_{l2}, \ldots, f_{lN}\} \). The instruction-router \( R \) dynamically generates \( N \) detailed and adaptive prompts based on the input instruction \( T \) and the instruction-aware VLM features \( F_{\text{VLM}} \):
\begin{equation}
    \{P_1, P_2, \ldots, P_N\} = R(T, F_{\text{VLM}})
\end{equation}
Here, \( R \) incorporates a feed-forward network (FFN) that projects the combined instruction \( T \) and visual features \( F_{\text{VLM}} \) into the embedding space for each sub-task. This dynamic prompt generation ensures that each \( P_i \) is tailored to its respective sub-task. These tailored prompts are then directed by the instruction router to their corresponding LLMs, enabling each LLM to focus on producing precise and contextually relevant outputs for its assigned sub-task. In our experiment, the task is divided into three subtasks: thematic interpretation, object identification, and analysis of object relationships. Each LLM \( f_{li} \) focuses on one sub-task, with its output expressed as:
\begin{equation}
    s_i = f_{li}(P_i, F_{\text{VLM}}), \quad i = 1, 2, 3
\end{equation}
The final caption \( S \) is obtained by aggregating the outputs of all LLMs:
\begin{equation}
    S = \text{Aggregate}(\{s_1, s_2, s_3\})
\end{equation}
By minimizing the loss of each LLM output against its respective subtask-specific ground truth, each LLM is guided to specialize in a narrow aspect of the task, reducing complexity and enhancing precision. This modular design, combined with dynamically adjusted prompts during training, ensures that the instruction router provides task-relevant guidance. \minorrevise{Generalization errors are computed based on the subtask-specific ground truth, ensuring that the evaluation aligns with each subtask's intended learning objective. This approach prevents misleading assessments that could arise from evaluating outputs against an overly broad reference.} Ultimately, this framework leverages the combined expertise of multiple LLMs to significantly improve caption generation performance.

}
\subsection{Model Tuning} \label{sec:tuning} {\color{black}

As illustrated in Fig.~\ref{fig_moe}, the tuning process of the proposed RS-MoE model is divided into two stages. Considering the complexity and diversity of information within remote sensing images, directly training our model can lead to performance degradation due to sparsity. This sparsity stems from the inadequate representation of remote sensing-specific features in models pre-trained on natural images, creating inefficiencies in capturing the intricate characteristics of remote sensing imagery, ultimately impacting model performance. To address this, we propose a two-stage training strategy. The first stage begins by independently training the VLM Encoder and LLM, ensuring that the encoder can extract rich and detailed visual features specific to remote sensing imagery while enhancing its capability to generate meaningful, context-aware feature representations. Once the VLM Encoder generates stable visual features, the second stage involves fine-tuning the language models in the MoE. In this stage, focus shifts to adjusting the MoE Block’s weights to produce more detailed captions tailored specifically for RSIC tasks, with each language model concentrating on its designated task.

\subsubsection{Stage I} In this stage, our primary focus is on tuning the weights of the VLM Encoder and the LLM Block to initialize model weights specifically designed for the RSIC task. Specifically, we freeze the Image Encoder and then train the VLM Encoder based on the extracted visual features and input instructions. This tuning enhances the encoder's ability to associate specific visual features with relevant instructional information, thereby improving its capability to generate meaningful and context-aware feature representations from remote sensing images. To significantly reduce the number of trainable parameters, we incorporate the LoRA training strategy during the VLM Encoder training stage. Firstly, within the self-attention layers, LoRA is specifically applied to the query and value projection layers. Secondly, LoRA is also implemented across all projection layers within the cross-attention layers, including those for the query, key, value, and output. Thirdly, LoRA is integrated into the feed-forward network components, ensuring a comprehensive application of this strategy throughout the model.

\revise{The LLM block receives output features from the VLM Encoder and input instructions to generate the final captions for remote sensing images,} which plays a crucial role in synthesizing the contextual data provided by the VLM Encoder into coherent and detailed textual descriptions. The fine-tuning process is focused on enhancing the model's ability to understand the complex visual patterns typical of remote sensing images and to translate them into accurate linguistic representations. By tuning the LLM Block, the model becomes more efficient at generating captions that are deeply aligned with the specific requirements of remote sensing tasks, including adapting to various landscapes, recognizing subtle environmental changes, and describing intricate relationships between different geographical features in the images. Similarly, we employ the LoRA technique to reduce the number of trainable parameters, thereby enhancing the model's efficiency and manageability during the training process. LoRA is specifically applied to the query and value layers within the attention module, as well as to the feed-forward networks. This optimization not only streamlines the training process but also sharpens the focus of the attention mechanisms, improving the overall processing and responsiveness of the model.

\subsubsection{Stage II} In this stage, our focus shifts to fine-tuning \revise{the weights of the LLMs within the MoE Block} to produce more detailed captions specifically for remote sensing image captioning tasks. Tuning the MoE Block involves detailed adjustments to the interaction mechanisms between the distinct LLMs to ensure that each LLM can effectively respond to its specific prompt by extracting and interpreting nuanced visual cues. This fine-tuning facilitates the development of a deep, segmented understanding of the image, allowing each expert within the MoE to generate a part of the caption that reflects its specialized knowledge of the image's theme, objects, or relationships. As these parts come together, they form a comprehensive and detailed caption that captures both the breadth and depth of the remote sensing imagery.

Specifically, we freeze the Image Encoder and the VLM Encoder, where the weights of the VLM Encoder are copied from those trained in Stage I. The MoE Block uses the Instruction Router to convert the instruction into specific prompts, which are then combined with the features extracted by the VLM Encoder related to the remote sensing images and input into different LLMs. By tuning the MoE, which principally involves tuning each of the sub-LLM Blocks within it, we initialize the weights of each LLM Block using the weights from the LLM Blocks that were trained in Stage I. During the tuning process, we continue to employ the LoRA training strategy to further enhance the models' efficiency and effectiveness. This approach not only streamlines the training by reducing the number of parameters that need to be tuned but also assists in fine-tuning the model's ability to concentrate on specific features of the input data, thereby enhancing the overall accuracy and responsiveness of our system.

}

\section{Experiments}

\subsection{Datasets}

\subsubsection{RSICap} This dataset is built on the DOTA object detection dataset, which contains 2,585 high-quality remote sensing image-text pairs for training vision-language models on remote sensing images. The images are randomly chosen from the DOTA dataset and resized to 512 $\times$ 512 pixels. The text annotations are manually labeled by five experts in the remote sensing field. To evaluate the VLM model on this dataset, the authors further constructed an evaluation set, RSIEval, that consists of 1,000 512x512 remote sensing images with manual annotations.

\subsubsection{UCM-Captions} This dataset is derived from the UC Merced Land Use Dataset. It consists of 2,100 remote sensing images, categorized into 21 land use classes, each containing 100 images. Each image measures 256 $\times$ 256 pixels, with a pixel resolution of 0.3048 meters. The image captions were manually extracted from large-scale images provided by the United States Geological Survey's National Map Urban Area Imagery. Each image is described by five sentences, which are relatively fixed in structure and simple in their descriptions

\subsubsection{Sydney-Captions} This dataset is constructed from the Sydney Data Set. It comprises 613 remote sensing images divided into 7 land use categories. Each image measures 500 $\times$ 500 pixels, with a pixel resolution of 0.5 meters. Each image is described in five sentences.

\subsubsection{RSICD} This dataset contains 10,921 remote sensing images collected from four platforms: Google Earth, Baidu Maps, MapABC, and Tianditu. The images are uniformly resized to 224 $\times$ 224 pixels, although they come in various resolutions. The captions for these images were created by volunteers who possess experience in annotation and knowledge of remote sensing. For each image, the original caption varies from one to five sentences. The authors duplicated the captions that contained fewer than five sentences to ensure that each image had five distinct sentences in its caption.

{\color{black}

\subsubsection{VRSBench} This dataset consists of 29,614 images, each paired with a human-verified, detailed caption. These captions include abstract image attributes, such as source, resolution, and scene type, alongside object-specific details like quantity, color, shape, and relative positioning. They emphasize clear, observable features, excluding ambiguous elements, and may describe additional objects, such as buildings, roads, and trees, beyond those in the source datasets. Each caption typically comprises 3-7 sentences, averaging 54 words.
}

\begin{table*}[!h]
    \centering
    \caption{Results of our model and state-of-the-art models on the RSIEval dataset for RSIC Task.}
    \resizebox{0.9\textwidth}{!}{
    \begin{tabular}{lccccccc}
         \toprule
         Method & BLEU-1 & BLEU-2 & BLEU-3 & BLEU-4 & METEOR & ROUGE\_L & CIDEr \\
         \midrule
         BLIP2-13B~\cite{li2023blip} & 54.51 & 42.42 & 34.64 & 28.17 & 27.54 & 24.38 & 105.51\\
         MiniGPT4-13B~\cite{zhu2023minigpt} & 68.49 & 59.31 & 43.22 & 39.78 & 31.70 & 30.95 & 120.33\\
         InstructBLIP-13B~\cite{dai2023instructblip} & 63.71 & 50.76 & 49.35 & 40.50 & 30.58 & 27.15 & 121.91\\
         RSGPT-13B~\cite{hu2023rsgpt} & 77.05 & 62.18 & 48.25 & 40.34 & 37.41 & 33.26 & 149.32\\
         {Qwen2-VL-7B~\cite{yang2024qwen2}} & {78.66} & {60.98} & {47.07} & {39.24} & {40.05} & {32.63} & {124.36} \\
         {LLaVA-NeXT-7B~\cite{li2024llava}} & {81.72} & {63.99} & {49.31} & {40.38} & {39.59} & {34.62} & {128.27} \\
         \midrule
         {\modelname-1B} & {57.36} & {42.25} & {22.14} & {20.54} & {32.36} & {25.98} & {109.37}\\
         {\modelname-3B} & {65.11} & {52.44} & {28.78} & {26.86} & {40.04} & {27.62} & {120.20} \\
         \modelname-7B & \textbf{ 82.13} & \textbf{65.44} & \textbf{51.93} & \textbf{42.55} & \textbf{40.28} & \textbf{35.72} & \textbf{158.36} \\
         \bottomrule
    \end{tabular}
    }
    \label{tab_res_rsieval}
\end{table*}

\begin{figure*}
    \centering   
    \includegraphics[width=\linewidth]{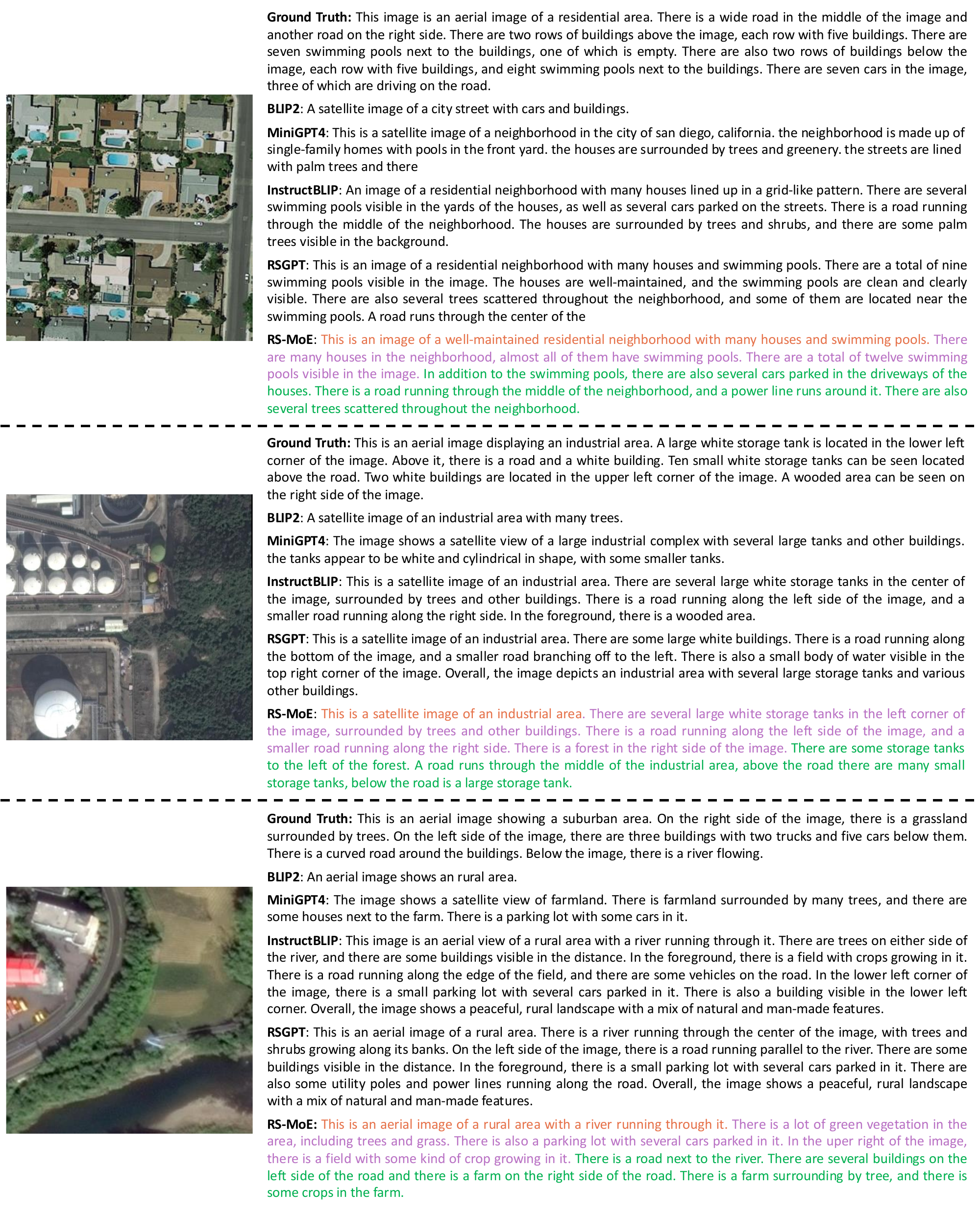}
    \caption{{Qualitative results of image captions generated by our~\modelname~model and the SOTA model for three randomly selected remote sensing images from the RSEval dataset. In captions generated by our model, different colors indicate distinct aspects of the image: orange for the overall theme, purple for objects within the scene, and green for spatial relationships between objects.}}
    \label{fig_res_rsieval}
\end{figure*}

\subsection{Evaluation Metrics}
\textit{BiLingual Evaluation Understudy (BLEU)} operates by comparing the text generated by the model to reference texts that are considered correct. BLEU examines the presence and frequency of n-grams in the generated text that appears in the reference text and calculates a score based on the precision of these n-grams. We tested BLEU scores for n-grams of sizes 1, 2, 3, and 4.

\textit{Recall-Oriented Understudy for Gisting Evaluation (ROUGE-L)} is a metric designed to evaluate the quality of summaries by measuring the longest common subsequence (LCS) between the generated text and the reference text. In the context of image captioning, ROUGE-L assesses the extent to which the content of a generated caption overlaps with that of a set of reference captions. This metric is recall-oriented, focusing on the coverage provided by the generated caption of the reference captions.

\textit{Metric for Evaluation of Translation with Explicit Ordering (METEOR)} is an evaluation metric that considers not only the precision but also the recall of matched words between the generated text and reference texts, thus balancing accuracy and coverage. It also accounts for synonymy and stemming, allowing for a more nuanced comparison of content.

\textit{Consensus-based Image Description Evaluation (CIDEr)} is specifically developed for scoring image captions by comparing them to reference captions written by humans. It measures the consensus between a candidate caption and the set of reference captions. This metric evaluates the similarity of n-grams between the generated and reference captions, weighting more frequent n-grams higher to reflect their importance in the descriptions.

\subsection{Experimental Details}

We trained our model on 3 NVIDIA A100 GPUs, each with 80GB of memory. {We selected three distinct versions of LLMs for use in our MoE model: Llama-3.2-1B~\cite{dubey2024llama}, Llama-3.2-3B~\cite{dubey2024llama}, and Vicuna-7B~\cite{vicuna2023}.} Each LLM within the MoE Block was individually fine-tuned for only 5 epochs.  During the training process, the learning rate was initially set to $1e-4$, and a warm-up phase for the learning rate was applied for one epoch. We used the AdamW optimizer with parameters set to $\beta_1$ = 0.9, $\beta_2$ = 0.999, and a weight decay of 0.05, alongside adopting a cosine decay schedule for the reduction of the learning rate.

\subsection{Results}

\begin{table*}[!h]
    \centering
    \caption{Results of our model and state-of-the-art models on the UCM-Captions dataset for RSIC Task.}
    \resizebox{0.9\textwidth}{!}{
    \begin{tabular}{lccccccc}
         \toprule
         Method & BLEU-1 & BLEU-2 & BLEU-3 & BLEU-4 & METEOR & ROUGE\_L & CIDEr \\
         \midrule
         VLAD + RNN ~\cite{rsicd} & 63.11 & 51.93 & 46.06 & 42.09 & 29.71 & 58.78 & 200.66 \\
         VLAD + LSTM ~\cite{rsicd} & 70.16 & 60.85 & 54.96 & 50.30 & 34.64 & 65.20 & 231.31 \\
         mRNN ~\cite{ucm_sydney_caption} & 60.10 & 50.70 & 32.80 & 20.80 & 19.30 & - & 214.00 \\
         mLSTM ~\cite{ucm_sydney_caption} & 63.50 & 53.20 & 37.50 & 21.30 & 20.30 & - & 222.50 \\
         mGRU ~\cite{mgru} & 42.56 & 29.99 & 22.91 & 17.98 & 19.41 & 37.97 & 124.82 \\
         mGRU embedword ~\cite{mgru} & 75.74 & 69.83 & 64.51 & 59.98 & 36.85 & 66.74 & 279.24 \\
         CSMLF ~\cite{csmlf} & 37.71 & 14.85 & 7.63 & 5.05 & 9.44 & 29.86 & 13.51 \\
         SAA ~\cite{sound} & 79.62 & 74.01 & 69.09 & 64.77 & 38.59 & 69.42 & 294.51 \\
         Soft-attention ~\cite{softattention} & 74.54 & 65.45 & 58.55 & 52.50 & 38.86 & 72.37 & 261.24 \\
         Hard-attention ~\cite{softattention} & 81.57 & 73.12 & 67.02 & 61.82 & 42.63 & 76.98 & 299.47 \\
         SD-RSIC ~\cite{sd-rsic} & 74.80 & 66.40 & 59.80 & 53.80 & 39.00 & 69.50 & 213.20 \\
         RTRMN (semantic) ~\cite{rtrmn} & 55.26 & 45.15 & 39.62 & 35.87 & 25.98 & 55.38 & 180.25 \\
         RTRMN (statistical) ~\cite{rtrmn} & 80.28 & 73.22 & 68.21 & 63.93 & 42.58 & 77.26 & 312.70 \\
         SVM-D BOW ~\cite{svm-d} & 76.35 & 66.64 & 58.69 & 51.95 & 36.54 & 68.01 & 271.42 \\
         SVM-D CONC ~\cite{svm-d} & 76.53 & 69.47 & 64.17 & 59.42 & 37.02 & 68.77 & 292.28 \\
         Post-processing ~\cite{post-processing} & 79.73 & 72.98 & 67.44 & 62.62 & 40.80 & 74.06 & 309.64 \\
         RSGPT-13B ~\cite{hu2023rsgpt} & 86.12 & 79.14 & 72.31 & 65.74 & 42.21 & {78.34} & 333.23 \\
         {SkyEyeGPT-7B~\cite{zhan2024skyeyegpt}} & {90.71} & {85.69} & {\textbf{81.56}} & {\textbf{78.41}} & {46.24} & {79.49} & {236.75} \\
        {RS-LLaVA-13B~\cite{bazi2024rs}} & {90.00} & {84.88} & {80.30} & {76.03} & {49.21} & {\textbf{85.78}} & {355.61} \\
        {RS-CapRet-7B~\cite{silva2024large}} & {84.30} & {77.90} & {72.20} & {67.00} & {47.20} & {81.70} & {354.80} \\
         \midrule
         {\textbf{\modelname-7B~w/o finetuning}} & \textbf{94.81} & \textbf{87.09} & {79.57} & {72.34} & \textbf{66.97} & 62.74 & \textbf{396.46} \\
         \bottomrule
    \end{tabular}
    }
    \label{tab_res_ucm}
\end{table*}
\begin{figure*}[!h]
    \centering    \includegraphics[width=0.9\linewidth]{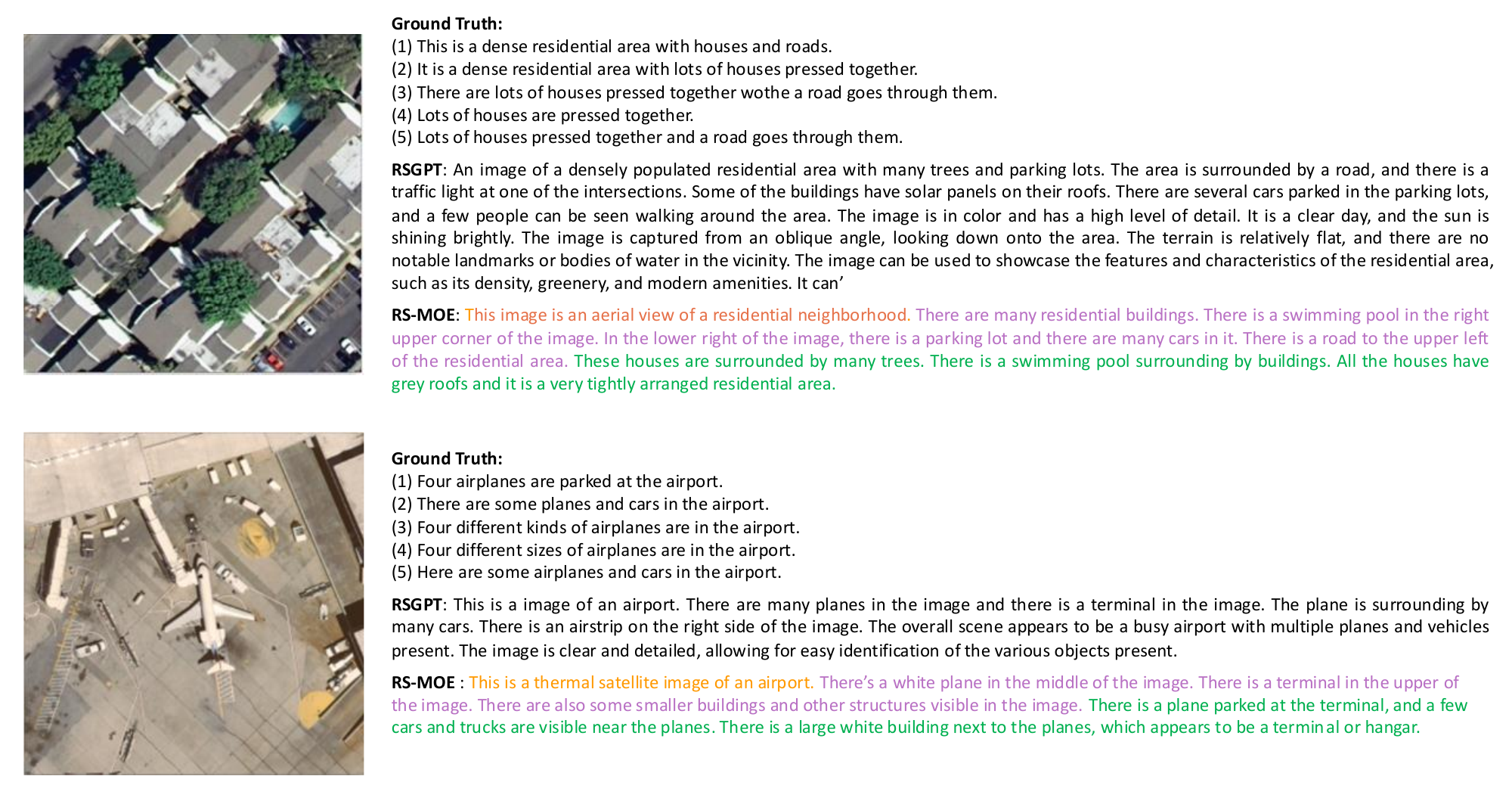}
    \caption{{Qualitative results of image captions generated by RSGPT and our~\modelname~model for two randomly selected remote sensing images from the UCM-Captions dataset. In captions generated by our model, different colors indicate distinct aspects of the image: orange for the overall theme, purple for objects within the scene, and green for spatial relationships between objects.}
    }
    \label{fig_res_ucm}
\end{figure*}

\begin{table*}[!h]
    \centering
    \caption{Results of our model and state-of-the-art models on the Sydney-Captions dataset for RSIC Task.}
    \resizebox{0.9\textwidth}{!}{
    \begin{tabular}{lccccccc}
         \toprule
         Method & BLEU-1 & BLEU-2 & BLEU-3 & BLEU-4 & METEOR & ROUGE\_L & CIDEr \\
         \midrule
         VLAD + RNN ~\cite{rsicd} & 56.58 & 45.14 & 38.07 & 32.79 & 26.72 & 52.71 & 93.72 \\
         VLAD + LSTM ~\cite{rsicd} & 49.13 & 34.12 & 27.60 & 23.14 & 19.30 & 42.01 & 91.64 \\
         mRNN ~\cite{ucm_sydney_caption} & 51.30 & 37.50 & 20.40 & 19.30 & 18.50 & - & 161.00 \\
         mLSTM ~\cite{ucm_sydney_caption} & 54.60 & 39.50 & 22.30 & 21.20 & 20.50 & - & 186.00 \\
         mGRU ~\cite{mgru} & 69.64 & 60.92 & 52.39 & 44.21 & 31.12 & 59.17 & 171.55 \\
         mGRU embedword ~\cite{mgru} & 68.85 & 60.03 & 51.81 & 44.29 & 30.36 & 57.47 & 168.94 \\
         CSMLF ~\cite{csmlf} & 59.98 & 45.83 & 38.69 & 34.33 & 24.75 & 50.18 & 75.55 \\
         SAA ~\cite{sound} & 68.82 & 60.73 & 52.94 & 45.39 & 30.49 & 58.20 & 170.52 \\
         Soft-attention ~\cite{softattention} & 73.22 & 66.74 & 62.23 & 58.20 & 39.42 & 71.27 & 249.93 \\
         Hard-attention ~\cite{softattention} & 75.91 & 66.10 & 58.89 & 52.58 & 38.98 & 71.89 & 218.19 \\
         SD-RSIC ~\cite{sd-rsic} & 72.40 & 62.10 & 53.20 & 45.10 & 34.20 & 63.60 & 139.50 \\
         SVM-D BOW ~\cite{svm-d} & 77.87 & 68.35 & 60.23 & 53.05 & 37.97 & 69.92 & 227.22 \\
         SVM-D CONC ~\cite{svm-d} & 75.47 & 67.11 & 59.70 & 53.08 & 36.43 & 67.46 & 222.22 \\
         Post-processing ~\cite{post-processing} & 78.37 & 69.85 & 63.22 & 57.17 & 39.49 & 71.06 & 255.53 \\
         RSGPT-13B~\cite{hu2023rsgpt}& 82.26 & 75.28 & 68.57 & 62.23 & 41.37 & 74.77 & 273.08 \\
         {RS-CapRet-7B~\cite{silva2024large}} & {78.70} & {70.00} & {62.80} & {56.40} & {38.80} & {70.70} & {239.20} \\
         \midrule
         {\textbf{\modelname-7B~w/o finetuning}} & \textbf{84.56} & \textbf{76.10} & \textbf{69.25} & \textbf{63.02} & \textbf{56.23} & \textbf{76.62} & \textbf{279.76} \\
         \bottomrule
    \end{tabular}
    }
    \label{tab_res_sydney}
\end{table*}

\begin{figure*}[!h]
    \centering \includegraphics[width=0.9\linewidth]{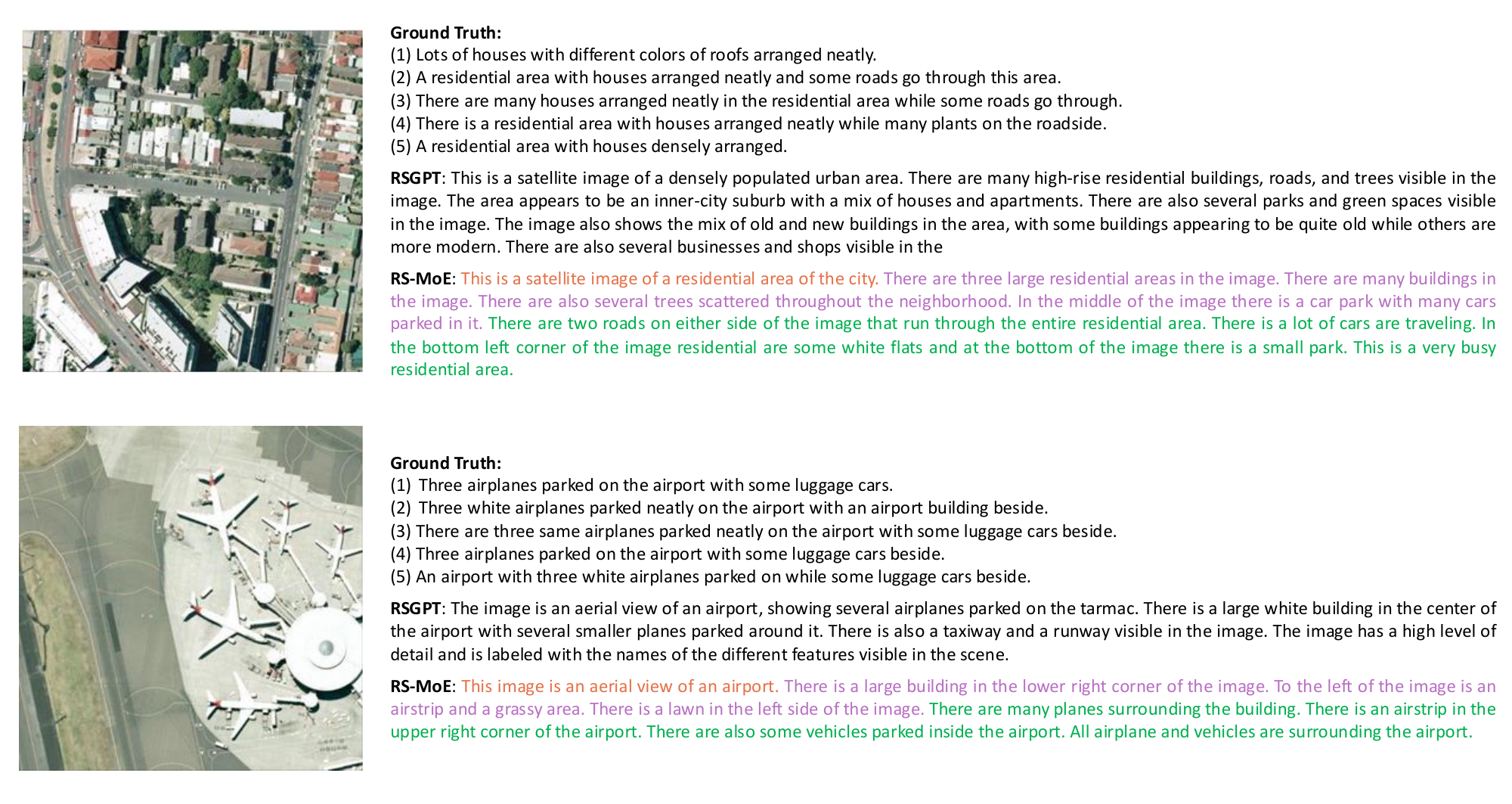}
    \caption{{Qualitative results of image captions generated by RSGPT and our~\modelname~model for two randomly selected remote sensing images from the Sydney-Captions dataset. In captions generated by our model, different colors indicate distinct aspects of the image: orange for the overall theme, purple for objects within the scene, and green for spatial relationships between objects.}
    }
    \label{fig_res_sydney}
\end{figure*}

\begin{table*}[!t]
    \centering
    \caption{Results of our model and state-of-the-art models on the RSICD dataset for RSIC Task.}
    \resizebox{0.9\textwidth}{!}{
    \begin{tabular}{lccccccc}
         \toprule
         Method & BLEU-1 & BLEU-2 & BLEU-3 & BLEU-4 & METEOR & ROUGE\_L & CIDEr \\
         \midrule
         VLAD + RNN ~\cite{rsicd} & 49.38 & 30.91 & 22.09 & 16.77 & 19.96 & 42.42 & 103.92 \\
         VLAD + LSTM ~\cite{rsicd} & 50.04 & 31.95 & 23.19 & 17.78 & 20.46 & 43.34 & 118.01 \\
         mRNN ~\cite{ucm_sydney_caption} & 45.58 & 28.25 & 18.09 & 12.13 & 15.69 & 31.26 & 19.15 \\
         mLSTM ~\cite{ucm_sydney_caption} & 50.57 & 32.42 & 23.29 & 17.46 & 17.84 & 35.02 & 31.61 \\
         mGRU ~\cite{mgru} & 42.56 &29.99 & 22.91 & 17.98 & 19.41 & 37.97 & 124.82 \\
         mGRU embedword ~\cite{mgru} & 60.94 & 46.24 & 36.80 & 29.81 & 26.14 & 48.20 & {159.54} \\
         CSMLF ~\cite{csmlf} & 57.59 & 38.59 & 28.32 & 22.17 & 21.28 & 44.55 & 52.97 \\
         SAA ~\cite{sound} & 59.35 & 45.11 & 35.29 & 28.08 & 26.11 & 49.57 & 132.35 \\
         Soft-attention ~\cite{softattention} & 65.13 & 49.04 & 39.00 & 32.30 & 26.39 & 49.69 & 90.58 \\
         SD-RSIC ~\cite{sd-rsic} & 64.50 & 47.10 & 36.40 & 29.40 & 24.90 & 51.90 & 77.50 \\
         RTRMN (semantic) ~\cite{rtrmn} & 62.01 & 46.23 & 36.44 & 29.71 & 28.29 & {55.39} & 151.46 \\
         RTRMN (statistical) ~\cite{rtrmn} & 61.02 & 45.14 & 35.35 & 28.59 & 27.51 & 54.52 & 148.20 \\
         SVM-D BOW ~\cite{svm-d} & 61.12 & 42.77 & 31.53 & 24.11 & 23.03 & 45.88 & 68.25 \\
         SVM-D CONC ~\cite{svm-d} & 59.99 & 43.47 & 33.55 & 26.89 & 22.99 & 45.57 & 68.54 \\
         MLAT ~\cite{mlat} & 66.90 & 51.13 & 41.14 & 34.21 & 27.31 & 50.57 & 94.27 \\
         Post-processing ~\cite{post-processing} & 62.90 & 45.99 & 35.68 & 28.68 & 25.30 & 47.34 & 75.56 \\
         RSGPT-13B ~\cite{hu2023rsgpt}& 70.32 & 54.23 & 44.02 & 36.83 & 30.10  & 53.34 & 102.94 \\
         {RS-CapRet-7B~\cite{silva2024large}} & {72.00} & {59.90} & {50.60} & {43.30} & {37.00} & {\textbf{63.30}} & {\textbf{250.20}} \\
         \midrule
         {\textbf{\modelname-7B~w/o finetuning}} & \textbf{78.61} & \textbf{63.60} & \textbf{51.44} & \textbf{43.35} & \textbf{39.30}  & {60.77} & {109.46} \\
         \bottomrule
    \end{tabular}
    }
    \label{tab_res_rsicd}
\end{table*}

\begin{figure*}[!h]
    \centering    \includegraphics[width=0.9\linewidth]{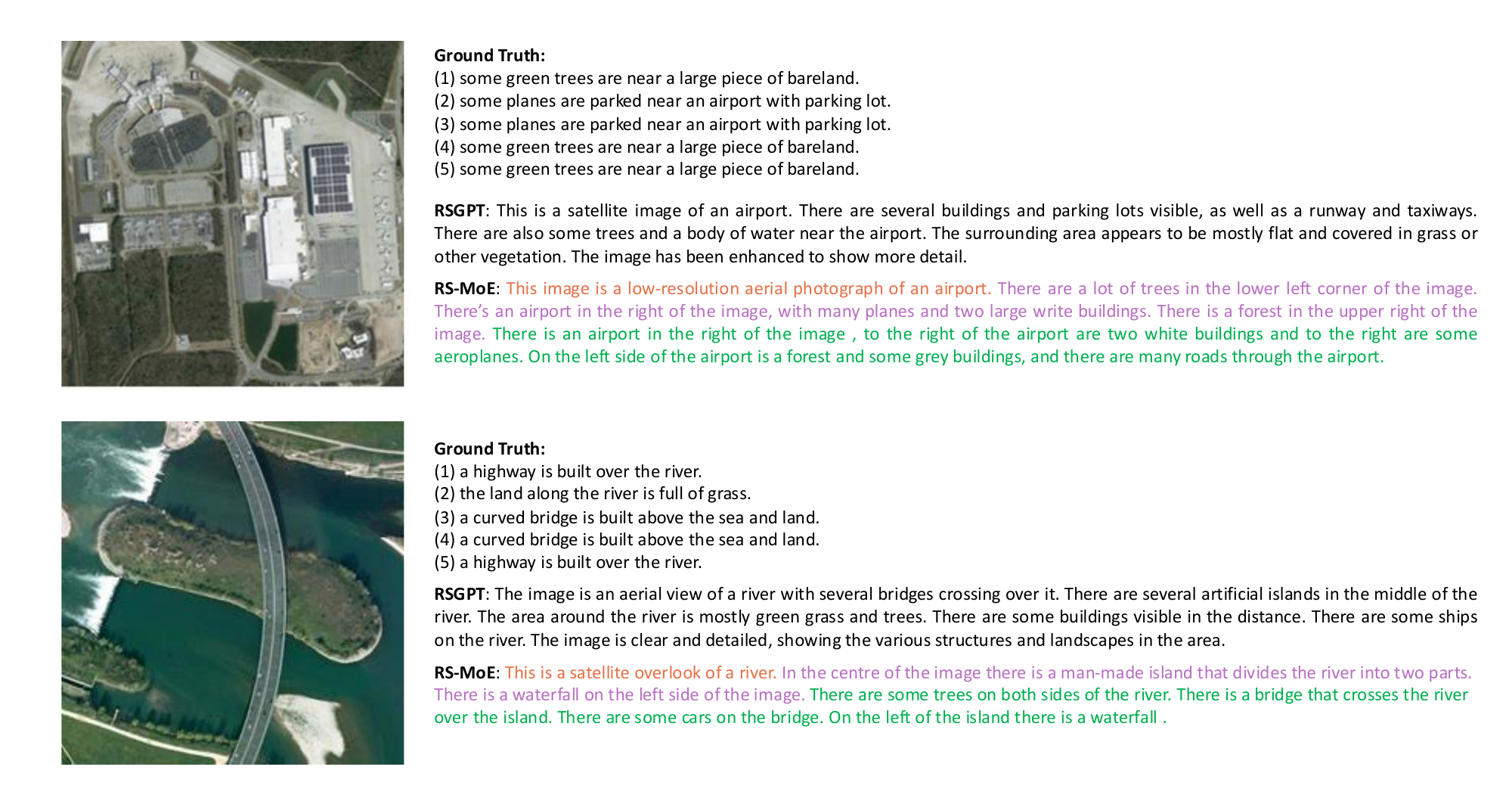}
    \caption{{Qualitative results of image captions generated by RSGPT and our~\modelname~model for two randomly selected remote sensing images from the RSICD dataset. In captions generated by our model, different colors indicate distinct aspects of the image: orange for the overall theme, purple for objects within the scene, and green for spatial relationships between objects.}}
    \label{fig_res_rsicd}
\end{figure*}

\begin{table*}[!h] \color{black}
    \centering
    \caption{Results of our model and state-of-the-art models on the VRSBench dataset for RSIC Task.}
    \resizebox{1.0\textwidth}{!}{
    \begin{tabular}{lccccccccc}
         \toprule
         Method & BLEU-1 & BLEU-2 & BLEU-3 & BLEU-4 & METEOR & ROUGE\_L & CIDEr & CHAIR\\
         \midrule
         GeoChat w/o ft~\cite{kuckreja2023geochat} & 13.9 &  6.6 & 3.0 & 1.4 &  7.8 & 13.2 & 0.4 & 0.42 \\
          GPT-4V~\cite{achiam2023gpt} & 37.2 &  22.5 &  13.7 & 8.6 &  20.9 & 30.1 & 19.1 & \textbf{0.83} \\
         MiniGPT-v2~\cite{chen2023minigpt} & 36.8 & 22.4 & 13.9 & 8.7 & 17.1 & 30.8 & 21.4 & 0.73\\
         LLaVA-1.5~\cite{liu2024improved} & 48.1 & 31.5 & 21.2 & 14.7 & 21.9 & 36.9 & 33.9 & 0.78\\
         GeoChat~\cite{kuckreja2023geochat} & 46.7& 30.2 & 20.1 & 13.8 & 21.1 & 35.2 & 28.2 & 0.77\\
         Mini-Gemini~\cite{li2024mini}  & 47.6 & 31.1 & 20.9 & 14.3 & 21.5 & 36.8 & 33.5 & 0.77\\
        \modelname & \textbf{55.3} & \textbf{34.9} & \textbf{25.5} & \textbf{17.2} & \textbf{25.1} & \textbf{42.8} & \textbf{40.5} & 0.81\\
         \bottomrule
    \end{tabular}
    }
    \label{tab_VRSBench}
\end{table*}

\subsubsection{Results on RSICap} {\color{black}

We first evaluated the proposed RS-MoE model against state-of-the-art VLMs on the RSICap dataset for the Remote Sensing Image Captioning task. Since these VLMs were initially designed for natural images, we fine-tuned each model according to the training strategies specified in their respective papers to ensure optimal adaptation for remote sensing tasks. To ensure a fair comparison, all models were fine-tuned for 5 epochs, aligning with the training regime of our RS-MoE. Subsequently, we tested all models, including three versions of RS-MoE, on the RSIEval dataset to assess their captioning accuracy. The results, displayed in Table~\ref{tab_res_rsieval}, demonstrate that our RS-MoE-7B model outperformed other models across all seven evaluation metrics, including four BLEU score variants, as well as METEOR, ROUGE-L, and CIDEr scores.

Notably, our two lightweight variants, RS-MoE-1B and RS-MoE-3B, achieved performance comparable to the larger 7B and 13B state-of-the-art models and even outperformed them on certain metrics. For instance, RS-MoE-1B exceeded BLIP2-13B in BLEU-1, METEOR, ROUGE-L, and CIDEr scores. Additionally, RS-MoE-1B not only surpassed BLIP2-13B but also outperformed MiniGPT4-13B and InstructBLIP-13B on the METEOR metric. These findings highlight the remarkable efficiency of our smaller models, as RS-MoE-1B and RS-MoE-3B deliver competitive performance while requiring significantly fewer parameters than their larger counterparts. Such compact architectures are particularly advantageous for resource-constrained environments, as they offer reduced computational overhead and are ideal for deployment in practical settings with limited memory and processing power.

Overall, the RS-MoE model demonstrates exceptional efficiency and accuracy in addressing the RSIC task. By leveraging its innovative architecture, including the MoE framework and specialized training strategy, the model effectively captures and interprets the complex visual information inherent in remote sensing images. This allows for the generation of precise, contextually relevant captions. Thus, our model not only excels in terms of performance but also proves highly practical for real-world applications, especially in environments where computational resources are limited.

To more intuitively compare the captioning performance of our model with other models for given remote sensing images, we randomly selected three images from the RSIEval dataset and listed the captions generated by each model in Figure~\ref{fig_res_rsieval}. These three images represent three distinct scenes: urban streets, industrial areas, and rural regions. By comparing the results of the generated captions from different models, several key findings can be made. The BLIP2 model provides relatively simple image captions, focusing mainly on the primary objects in the remote sensing images. The MiniGPT4 and InstructBLIP models incorporate more details and contextual information in their descriptions, such as trees around houses, swimming pools, and roads. These details enrich the descriptions but can sometimes result in information redundancy. The RSGPT model describes the images in a more structured manner. However, some of its generated descriptions may lack accuracy, such as the number of swimming pools, and may include some subjective and hallucinatory sentences. Conversely, our RS-MoE model describes the images with clear logic from three perspectives: themes, objects present, and the relationships between objects. The RS-MoE model provides more accurate descriptions of the types, quantities, and spatial relationships of the objects present in images.

}

\subsubsection{Results on UCM-Captions}  {\color{black}

We further validated the performance of our model on the classic UCM-Captions dataset. To establish a comprehensive comparison, we selected sixteen deep learning models specifically designed for the RSIC task, representing state-of-the-art approaches in remote sensing image captioning prior to the advent of VLM models. Additionally, we evaluated recent state-of-the-art VLM models in the remote sensing domain, including RSGPT \cite{hu2023rsgpt}, SkyEyeGPT \cite{zhan2024skyeyegpt}, RS-LLaVA \cite{bazi2024rs}, and RS-CapRet \cite{silva2024large}. Notably, we evaluated the performance of the 7B version of RS-MoE, which was trained on the RSICap dataset \textit{without any additional fine-tuning} on the UCM-Captions dataset. The comparative results of these baseline models and our model are presented in Table~\ref{tab_res_ucm}. Remarkably, even without fine-tuning on the UCM-Captions dataset, our model achieved the highest performance across several metrics, including BLEU-1, BLEU-2, METEOR, and CIDEr. Specifically, our model achieved a METEOR score of 66.97, significantly surpassing the second-highest score of 49.21 obtained by the RS-LLaVA-13B model, further demonstrating the robustness and generalizability of RS-MoE across different datasets.

To further illustrate the effectiveness of our model in generating accurate and contextually aware captions, we selected two remote sensing images with distinct scenes from the test set of the UCM-Captions dataset. The captions generated by RSGPT and our model for these images are shown in Figure~\ref{fig_res_ucm}. As observed, both models successfully capture the overall themes of the images. However, for detailed descriptions, particularly in the first remote sensing image, the captions generated by RSGPT exhibit discrepancies with the actual scene. In contrast, the captions produced by RS-MoE model provide precise and detailed representations of both the objects and their spatial relationships within the remote sensing images, underscoring our model’s advantage in understanding complex spatial arrangements in diverse remote sensing imagery.
}

\subsubsection{Results on Sydney-Captions}  {\color{black}
To further assess our model's efficacy for the RSIC task, we extended the evaluation to the traditional Sydney-Captions dataset. In this experiment, sixteen deep learning models specifically designed for remote sensing image captioning, along with two recent VLM models, RSGPT \cite{hu2023rsgpt} and RS-CapRet \cite{silva2024large}, were selected as comparative benchmarks. Noted that we still employed the 7B variant of RS-MoE \textit{without any additional fine-tuning} on the Sydney-Captions dataset. The comparative performance of these baseline models and our model is detailed in Table~\ref{tab_res_sydney}. Analysis of the results demonstrates that, even without fine-tuning, our model consistently outperformed all competing models across all seven evaluation metrics.A particularly notable achievement is its performance on the METEOR metric, where it achieved a score of 56.23, surpassing the next best model's score of 41.37 by a significant margin of 15\%.

To further illustrate the captioning capabilities of our model in comparison to RSGPT, we randomly selected two remote sensing images from the test set of the Sydney-Captions dataset, as shown in Figure~\ref{fig_res_sydney}. These images depict scenes of a residential area and an airport. Figure~\ref{fig_res_sydney} shows that while both RSGPT and RS-MoE provide generally accurate descriptions of the overall themes and specific details, the RSGPT-generated captions lack the depth observed in our model’s outputs. Additionally, the last sentence generated by RSGPT for the first image remains incomplete, underscoring our model’s advantage in generating contextually descriptions.
}

\begin{table*}[!h] \color{black}
    \centering
    \caption{Comparison results of RS-MoE with and without the instruction router on the RSICap dataset for the RSIC task.}
    \label{table_ab_training}
    \resizebox{0.9\textwidth}{!}{
    \begin{tabular}{lccccccc}
         \toprule
         Training Strategy & BLEU-1 & BLEU-2 & BLEU-3 & BLEU-4 & METEOR & ROUGE\_L & CIDEr \\
         \midrule
         \modelname~w/o Instruction Router & 47.32 & 33.18 & 25.66 & 21.42 & 22.87 & 25.33 & 102.51\\
         \modelname~w/ Instruction Router & 82.13 & 65.44 & 51.93 & 42.55 & 40.28 & 35.72 & 158.36\\
         \bottomrule
    \end{tabular}
    }
    \label{tab_router}
\end{table*}

\begin{table*}[!h]
    \centering
    \caption{Comparison results of different numbers of LLMs in the MoE block on the RSICap dataset for the RSIC task.}
    \resizebox{0.9\textwidth}{!}{
    \begin{tabular}{cccccccc}
         \toprule
         Number of LLMs & BLEU-1 & BLEU-2 & BLEU-3 & BLEU-4 & METEOR & ROUGE\_L & CIDEr \\
         \midrule
         1 & 48.51 & 38.65 & 29.57 & 25.13 & 23.01 & 21.49 & 93.51\\
         2 & 64.52 & 51.32 & 40.43 & 33.42 & 30.61 & 28.27 & 124.41\\
         3 &\textbf{82.13} & \textbf{65.44} & \textbf{51.93} & \textbf{42.55} & \textbf{40.28} & \textbf{35.72} & \textbf{158.36}  \\
         4 & 77.36 & 60.38 & 48.91 & 39.26 & 37.94 & 32.95 & 147.24\\
         \bottomrule
    \end{tabular}
    }
    \label{tab_num_llm}
\end{table*}

\subsubsection{Results on RSICD}  {\color{black}

Finally, we validated the performance of our model on the widely used RSICD dataset for the RSIC task. Consistent with previous evaluations, we compared our model against sixteen deep learning models and two recent VLM models, RSGPT \cite{hu2023rsgpt} and RS-CapRet \cite{silva2024large}, with our RS-MoE-7B model evaluated \textit{without additional fine-tuning} on the RSICD dataset. The comparative results for these baseline models and our model are presented in Table~\ref{tab_res_rsicd}. As observed in Table~\ref{tab_res_rsicd}, even without fine-tuning, our model achieved the highest scores on five out of seven metrics, including BLEU-1, BLEU-2, BLEU-3, BLEU-4, and METEOR. This highlights the robustness and adaptability of our approach, underscoring its ability to perform well across diverse datasets without requiring dataset-specific adjustments.

To further illustrate the effectiveness of the generated captions, we selected two remote sensing images from the test set of the RSICD dataset and displayed the captioning results generated by RSGPT and our model in Figure~\ref{fig_res_rsicd}. It is evident from Figure~\ref{fig_res_rsicd} that, unlike RSGPT, our RS-MoE model accurately describes the spatial positions of objects within the images. This capability demonstrates RS-MoE’s enhanced ability to capture and articulate spatial relationships and arrangements, which are essential for a comprehensive understanding of remote sensing imagery. By providing detailed, contextually rich descriptions that reflect the actual scene, RS-MoE ensures that the generated captions are both informative and precise, thereby improving the overall interpretability.

}

{\color{black}
\subsubsection{Results on VRSBench}  
We further evaluate our method on the RSBench dataset, a benchmark specifically designed to test detailed captioning capabilities. In our experiments, we compare our method against several strong baselines, including GeoChat without finetuning~\cite{kuckreja2023geochat}, GPT-4V~\cite{achiam2023gpt}, MiniGPT-v2~\cite{chen2023minigpt}, LLaVA-1.5~\cite{liu2024improved}, GeoChat~\cite{kuckreja2023geochat}, and Mini-Gemini~\cite{li2024mini}. The results of these models are summarized in Table~\ref{tab_VRSBench}. As shown in the table, our method demonstrates outstanding performance across all major metrics. Notably, our method achieves a BLEU-1 score of 55.3, significantly surpassing Mini-Gemini's BLEU-1 score of 47.6. Similarly, our method achieves a BLEU-3 score of 25.5, outperforming Mini-Gemini's 20.9. Although GPT-4V achieves the highest CHAIR score of 0.83, our method achieves a competitive score of 0.81, substantially outperforming other baselines. These results clearly demonstrate the effectiveness of our method in generating detailed and descriptive captions that capture both abstract image attributes and object-specific details. The substantial improvements over state-of-the-art baselines on RSBench further highlight the robustness and adaptability of our method in addressing the unique challenges of vision-language modeling for detailed captioning tasks.
}

\section{Discussion}

\revise{In this section, we conduct a comprehensive analysis of the effectiveness and generalizability of the proposed RS-MoE model from two perspectives. Firstly, we verify the effectiveness of our model and training strategy through four ablation studies. These include examining the effect of the Instruction Router, varying the number of LLMs in the MoE block, using different types of LLMs, and evaluating the two-stage training strategy. Subsequently, we extend the application of our model to the Remote Sensing Visual Question Answering (RSVQA) task, testing its performance on two datasets, RSEval and RSIVQA, to assess its broader applicability and generalization capabilities.}

\subsection{Ablation Study}

{\color{black}
\subsubsection{Effect of Instruction Router}
To evaluate the effectiveness of the instruction router within the MoE Block, we conducted an ablation study comparing the model’s performance with and without the instruction router. For a fair comparison, both setups use three LLMs as expert models. When the instruction router is absent, the input instructions are directly passed to the LLMs without any dynamic prompt generation, and each LLM independently generates outputs.The results of the two models are shown in Table~\ref{tab_router}. It can be found that the model without the instruction router achieves a BLEU-1 score of 47.32, a METEOR score of 22.87, and a CIDEr score of 102.51. In contrast, incorporating the instruction router leads to significant improvements, with the model achieving a BLEU-1 score of 82.13, a METEOR score of 40.28, and a CIDEr score of 158.36. These results highlight the critical role of the instruction router in dynamically generating tailored prompts based on input instructions and visual features, effectively guiding the expert models to produce more precise and contextually appropriate descriptions. The substantial performance gains observed with the instruction router emphasize its importance in enabling the model to adapt to diverse input images and tasks. By leveraging Prompt Learning principles, the instruction router enhances the model’s ability to extract meaningful visual information and generate accurate, detailed descriptions, demonstrating its indispensability in the proposed framework.}

\begin{table*}[!h] \color{black}
    \centering
    \caption{Comparison results of different LLM model in the MoE block on the RSICap dataset for the RSIC task.}
    \resizebox{0.9\textwidth}{!}{
    \begin{tabular}{lccccccc}
         \toprule
         LLM in MoE Block & BLEU-1 & BLEU-2 & BLEU-3 & BLEU-4 & METEOR & ROUGE\_L & CIDEr \\
         \midrule
         Llama-3.2-1B~\cite{dubey2024llama} & 57.36 & 42.25 & 22.14 & 20.54 & 32.36 & 25.98 & 109.37\\
          StableLM-1.6B~\cite{bellagente2024stable} & 56.42 & 44.36 & 23.27 & 22.14 & 33.71 & 24.62 & 110.85\\
         Qwen-1.8B~\cite{bai2023qwen} & 59.35 & 47.28 & 27.52 & 22.95 & 34.62 & 25.38 & 113.98\\
         Phi2-2.7B~\cite{javaheripi2023phi} & 59.78 & 48.15 & 27.35 & 25.58 & 36.77 & 25.36 & 114.96\\
         Llama-3.2-3B~\cite{dubey2024llama} & 65.11& 52.44 & 28.78 & 26.86 & 40.04 & 27.62 & 120.20\\
         LLaMa-7B~\cite{touvron2023llama}  & 77.95 & 62.11 & 49.28 & 40.27 & 38.26 & 31.91 & 150.31\\
         Vicuna-7B\cite{vicuna2023} & \textbf{ 82.13} & \textbf{65.44} & \textbf{51.93} & \textbf{42.55} & \textbf{40.28} & \textbf{35.72} & \textbf{158.36} \\
         \bottomrule
    \end{tabular}
    }
    \label{tab_ab_llm}
\end{table*}
\begin{table*}[!h] \color{black}
    \centering
    \caption{Comparison results of different training strategy for our model on the RSICap dataset for the RSIC task.}
    \label{table_ab_training}
    \vspace{0.1cm}
    \resizebox{0.9\textwidth}{!}{
    \begin{tabular}{lccccccc}
         \toprule
         Training Strategy & BLEU-1 & BLEU-2 & BLEU-3 & BLEU-4 & METEOR & ROUGE\_L & CIDEr \\
         \midrule
         One-Stage & 50.39 & 33.44 & 24.37 & 20.86 & 27.49 & 25.36 & 98.21\\
         Two-Stage & 82.13 & 65.44 & 51.93 & 42.55 & 40.28 & 35.72 & 158.36\\
         \bottomrule
    \end{tabular}
    }
    \label{tab_twostage}
\end{table*}

\subsubsection{Effect of Different Numbers of LLMs in the MoE Block}

As described in Section~\ref{sec:model}, within the MoE-Block, we employ an instruction router to generate three detailed prompts, each directed to a distinct Large Language Model (LLM) to independently handle the overall theme, geographic objects, and relationships among objects in remote sensing images. In this section, we analyze the results obtained from using varying numbers of LLMs. Specifically, we trained versions of the RS-MoE model with one to four LLMs on the RSICap dataset, keeping the first stage constant while adjusting the number of LLMs in the second stage. When the number of LLMs is set to 1, the model reduces to a conventional single-LLM decoder VLM. With two LLMs, each LLM is tasked with describing either the image’s theme or specific information. For configurations with four LLMs, we assign one LLM each to address the image's theme, objects, absolute positions of objects, and relationships among objects.

We evaluated the performance of these models on the RSIC task using the RSIEval dataset, with detailed results presented in Table~\ref{tab_num_llm}. By comparing the results for configurations with more than one LLM to those with only a single LLM, we observe substantial performance improvements when the task is divided into subtasks. These results further validate the effectiveness of incorporating the idea of MoE into the remote sensing image captioning task. Additionally, comparing the results for models with 2, 3, and 4 LLMs, we found that using three LLMs yields the best performance. Consequently, in our experiments, we employ three LLMs to focus on generating captions for the image’s theme, included objects, and spatial relationships among objects, highlighting the advantage of the MoE framework for handling complex, multi-faceted tasks in remote sensing. This setup not only enhances the interpretability of the generated captions but also demonstrates the flexibility and scalability of our model when addressing intricate details in remote sensing imagery.

\begin{table*}[!h]
    \centering
    \caption{Results of our model and state-of-the-art models on the RSVQA test set of RSIEval for RSVQA task.}
    \resizebox{\textwidth}{!}{
    \begin{tabular}{lcccccccccccc}
        \toprule
        Method & Presence & Quantity & Color & Absolute pos. & Relative pos. & Area comp. & Road dir. & Image & Scene & Reasoning & Avg accuracy \\
        \midrule
        BLIP2 &       60.41& 26.02& 43.24&  7.69& 13.16& 58.14& 33.33& 74.42& 43.24& 47.50& 45.56\\
        MiniGPT4&     29.70&  9.76& 31.53&  1.54&  1.32& 16.28&  0.00& 34.88& 24.32& 17.50& 21.82\\
        InstructBLIP& 76.14& 21.95& 45.05& 12.31& 10.53& 69.77&  0.00& 81.40& 45.95& 57.50& 53.26\\
        RSGPT& 81.22 & \textbf{39.02} & 54.05 & 38.46 & 35.53 & 62.79 & \textbf{66.67} & \textbf{93.02} & 89.19 & 70.00 & 65.24 \\
        \modelname & \textbf{83.16} & 32.77 & \textbf{71.92} & \textbf{43.95} & \textbf{45.00} & \textbf{63.41} & 0.00 & 77.89 & \textbf{93.18} & \textbf{74.07} & \textbf{68.65} \\
        \bottomrule
    \end{tabular}
    }
    
    \label{tab_rsieval_rsvqa}
\end{table*}

\begin{figure*}[!h]
    \centering    \includegraphics[width=\linewidth]{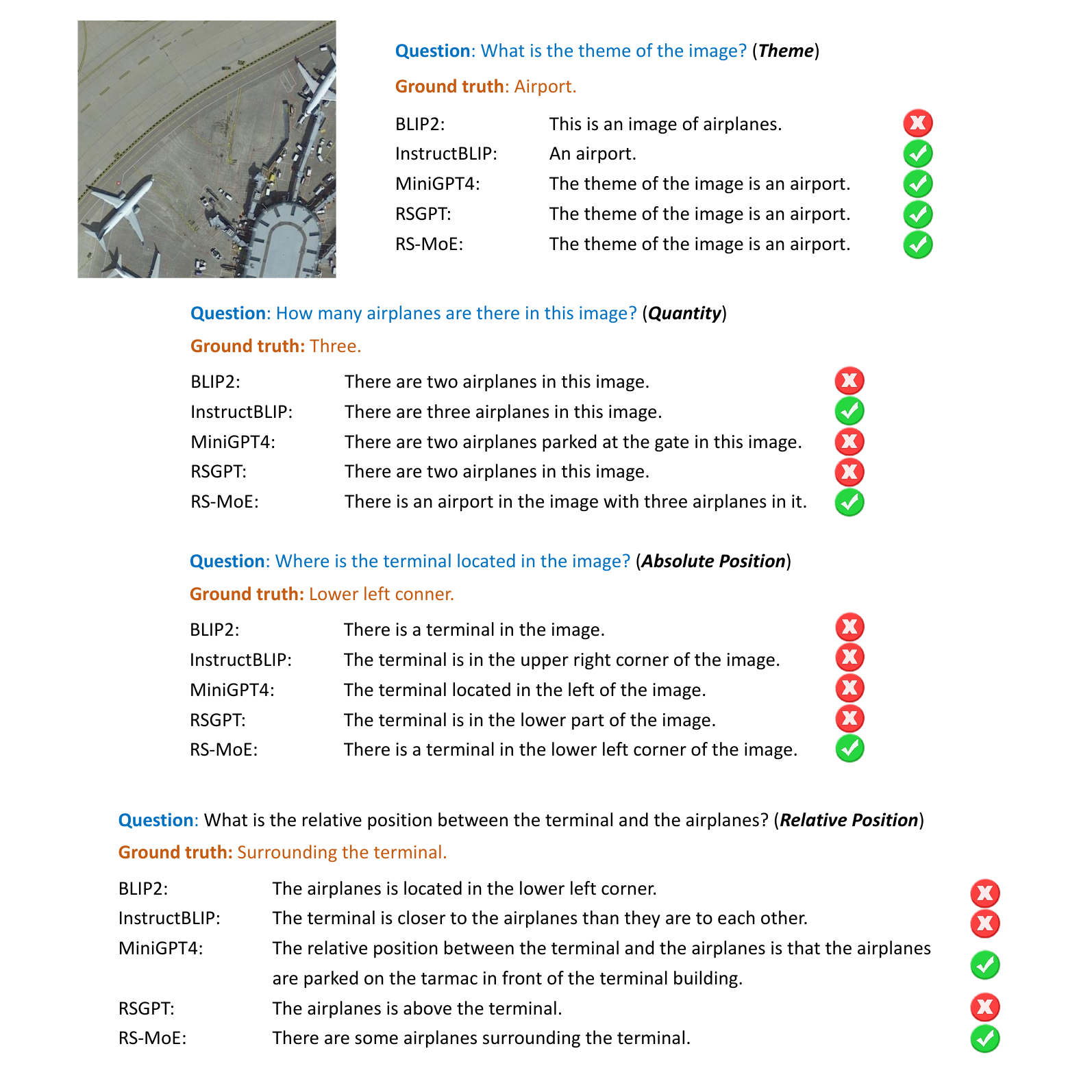}
    \caption{Qualitative results of visual question answering for remote sensing images by BLIP2, MiniGPT4, InstructBLIP, RSGPT, and our RS-MoE model on the RSVQA test set from the RSEval dataset.}
    \label{fig_res_vqa}
\end{figure*}

\subsubsection{Effect of Different LLM Models} {\color{black}

In this section, we investigate the effect of different lightweight LLMs within the MoE Block on remote sensing image captioning. Specifically, in addition to the original selection of Llama-3.2-1B, Llama-3.2-3B, and Vicuna-7B, we further experimented with four additional state-of-the-art LLMs from the computer vision field: Qwen-1.8B \cite{bai2023qwen}, StableLM-1.6B \cite{bellagente2024stable}, Phi2-2.7B \cite{javaheripi2023phi}, and LLama-7B \cite{touvron2023llama}. These models are designed to understand and generate human-like text across a wide range of topics, and we selected them to assess how different LLM configurations influence performance within our proposed~\modelname.

Each of the seven~\modelname model variants was trained as a baseline on the RSICap dataset and evaluated on the RSIEval dataset. The performance metrics are displayed in Table~\ref{tab_ab_llm}. As observed from the results, all selected lightweight LLMs under our MoE framework achieved reasonably satisfactory performance in remote sensing image captioning. This consistency across different LLMs demonstrates the robustness and adaptability of our proposed framework, indicating that even with various lightweight LLM configurations, the~\modelname model is able to maintain effective and contextually relevant caption generation for complex remote sensing imagery. This suggests that our MoE framework can be flexibly integrated with different LLMs, providing a versatile solution adaptable to the needs of various tasks and computational environments.}

\subsubsection{Effect of Two-stage Training Strategy} {\color{black}
We further evaluate the effectiveness of our proposed two-stage training strategy. As explained in Section~\ref{sec:tuning}, directly training our \modelname model can lead to performance degradation due to sparsity issues stemming from the distinct characteristics of remote sensing images compared to natural images. To validate this hypothesis, we implemented a baseline model trained using a single-stage approach, directly optimizing the model without employing the two-stage strategy. Comparative results for both training methods are shown in Table~\ref{table_ab_training}. The findings indicate that the two-stage training strategy significantly outperforms the single-stage approach, underscoring the advantages of this method. By incorporating the first training stage, the VLM encoder and LLM Block learn feature representations tailored to remote sensing imagery, gaining a foundational understanding of remote sensing-specific visual patterns and linguistic cues. This process establishes a robust basis for accurate feature extraction and preliminary captioning. This stage essentially fine-tunes the model's ability to interpret complex spatial arrangements and thematic elements unique to remote sensing imagery, elements often underrepresented in models pretrained on natural images. In the second stage, the model builds upon this solid foundation, requiring only minimal fine-tuning to adapt the MoE block. Consequently, each LLM in the MoE block can effectively focus on its designated task, whether it involves capturing thematic elements, identifying specific geographic objects, or interpreting spatial relationships among objects. This task-oriented fine-tuning results in captions that are not only precise but also provide a comprehensive view of the remote sensing scene, with detailed and contextually relevant descriptions. Overall, this two-stage training strategy represents a core innovation of our work, demonstrating how carefully designed training frameworks can achieve superior results in the specialized domain of remote sensing.
}

\begin{table*} \color{black}
\centering
\caption{Results of our model and state-of-the-art models on the RSIVQA dataset for RSVQA task.}
\label{tab_RSIVQA}
\resizebox{0.9\linewidth}{!}{
\begin{tabular}{lccccc}
\toprule
Model & Yes/No  & Number  & Others & Average Accuracy  & Overall Accuracy \\ 
\midrule
MAIN~\cite{zheng2021mutual}       & 92.82 & 56.71 & 54.50  & 68.01 & 77.39 \\
EasyToHard~\cite{yuan2022easy} & 95.49 & 49.03 & 63.65  & 69.39 & 79.70 \\
SHRNet~\cite{zhang2023spatial}     & 97.64 & 57.89 & 84.60  & 80.04 & 84.46 \\ 
RSAdapter~\cite{wang2024rsadapter}  &  97.90 & 62.64 & 92.47& 84.34 & 87.10 \\ 
\modelname  & \textbf{98.01} & \textbf{75.46}    & \textbf{92.88} & \textbf{88.78}    & \textbf{89.12} \\ 
\bottomrule
\end{tabular}
}
\end{table*}

\subsection{Extension on the RSVQA Task}
To verify the generalization of the proposed model, we further evaluated its performance on the visual question answering (VQA) task for remote sensing images. Specifically, we used the RSEval dataset's 10 categories of questions, including presence, quantity, color, absolute position, relative position, area comparison, road direction, image, scene, and reasoning. We queried BLIP2, MiniGPT4, InstructBLIP, RSGPT, and our RS-MoE model, all of which were trained on the RSICap dataset, and calculated the average accuracy of their answers for each category. As shown in Table~\ref{tab_rsieval_rsvqa}, our model achieved the highest performance in seven out of the ten categories. To visually demonstrate the results on RSVQA, we randomly selected a remote sensing image from the dataset and displayed the answers provided by these models in four aspects: Theme, Quantity, Absolute Position, and Relative Position. It was observed that all models were able to answer simple Theme questions accurately. However, as the complexity of the questions increased, especially for those related to spatial relationships, the baseline models' performance deteriorated, while our model continued to provide accurate answers. This indicates that our model excels in handling complex questions, demonstrating its superior capability in understanding and reasoning about spatial relationships in remote sensing images.

\revise{To evaluate the effectiveness of our model further, we conducted experiments on the RSIVQA dataset, a benchmark for the visual question answering (VQA) task in remote sensing. The RSIVQA dataset includes three question categories: Yes/No, Number, and Others, with the final results summarized in Table~\ref{tab_RSIVQA}. Our model is compared against state-of-the-art methods, including MAIN~\cite{zheng2021mutual}, EasyToHard~\cite{yuan2022easy}, SHRNet~\cite{zhang2023spatial}, and RSAdapter~\cite{wang2024rsadapter}. As shown in the table, our model achieves the best performance across all metrics, with an overall accuracy of 89.12\% and an average accuracy of 88.78\%. Notably, for the challenging Number category, our model achieves a significant improvement, with an accuracy of 75.46\%, outperforming the previous best model, RSAdapter, by 12.82\%. Similarly, in the Yes/No and Others categories, our model attains accuracies of 98.01\% and 92.88\%, respectively, setting new benchmarks in both cases. These results highlight the superior generalization ability of our model in handling diverse and complex question types in the RSIVQA dataset. Its exceptional performance, particularly in the Number and Others categories, underscores its capability to reason about numerical quantities and contextual relationships in remote sensing imagery. This demonstrates the robustness and adaptability of our approach in addressing the unique challenges posed by VQA tasks in remote sensing.
}

\section{Conclusion}

\revise{This paper introduces RS-MoE, a pioneering Mixture of Experts (MoE)-based Vision-Language Model (VLM) specifically developed for remote sensing image captioning. RS-MoE consists of four core modules: the Image Encoder, VLM Encoder, LLM Block, and MoE Block. At the heart of the framework, the MoE Block leverages an Instruction Router to dynamically generate task-specific prompts, guiding each Large Language Model (LLM) in subsequent layers to focus on distinct aspects of captioning—theme comprehension, object identification, and relationship interpretation. This design allows the model to capture intricate semantic features unique to remote sensing imagery, enabling the generation of detailed, contextually accurate captions. Additionally, we propose a two-stage training strategy to address performance degradation from sparsity issues inherent in MoE-based methods and enhance training efficiency. Our extensive experiments show that, even after fine-tuning on only one dataset, RS-MoE achieves state-of-the-art performance across five RSIC datasets and demonstrates robust generalization capabilities on two RSVQA datasets. Notably, our lightweight RS-MoE-1B variant achieves performance comparable to larger 13B VLMs, underscoring the model’s efficiency in remote sensing applications. Thus, RS-MoE not only excels in performance but also proves practical for real-world applications, particularly in resource-constrained environments where computational efficiency is essential.}

\revise{While RS-MoE demonstrates significant advantages, there are certain limitations that warrant further exploration. The reliance on predefined sub-tasks for LLM specialization may limit adaptability to highly complex or novel remote sensing scenarios. Additionally, the two-stage training strategy, while effective, introduces additional computational overhead during the pretraining phase. Future work could explore adaptive task routing mechanisms that allow for greater flexibility and generalization to new tasks. Furthermore, integrating multi-modal data sources, such as radar and hyperspectral imagery, could expand the applicability of RS-MoE to a broader range of remote sensing challenges.}






{\small
\bibliographystyle{IEEEtran}
\bibliography{bibliography}
}

\vspace{-10pt}
\begin{IEEEbiography}[{\includegraphics[width=1in,height=1.25in,clip,keepaspectratio]{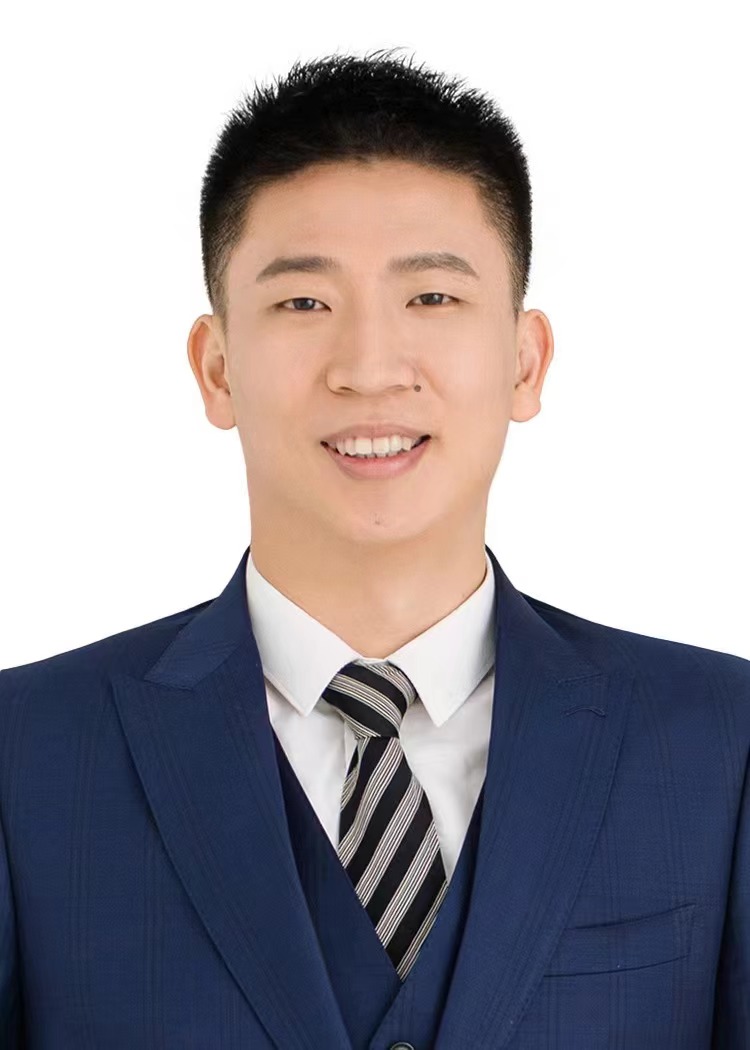}}]{Hui Lin} received the Bachelor's degree in Space Information and Digital Technology from Wuhan University, Wuhan, China, in 2012, and the Ph.D. degree in Cartography and Geographic Information Systems from the Institute of Remote Sensing and Digital Earth, Chinese Academy of Sciences, in 2017. He is currently a Senior Engineer in China Academy of Electronics and Information Technology, Beijing, China. His research interests include multimodal data fusion, deep learning, and remote sensing applications.
\end{IEEEbiography}

\begin{IEEEbiography}[{\includegraphics[width=1in,height=1.25in,clip,keepaspectratio]{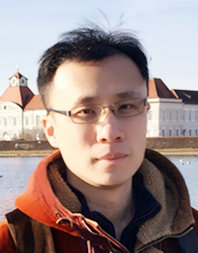}}]{Danfeng Hong}  (Senior Member, IEEE) received the Dr. -Ing degree (summa cum laude) from the Signal Processing in Earth Observation (SiPEO), Technical University of Munich (TUM), Munich, Germany, in 2019. 

Since 2022, he has been a Full Professor with the Aerospace Information Research Institute, Chinese Academy of Sciences. His research interests include Artificial Intelligence, Multimodal Big Data, Foundation Models, and Earth Observation. 

Dr. Hong serves as an Associate Editor for the IEEE Transactions on Image Processing (TIP) and the IEEE Transactions on Geoscience and Remote Sensing (TGRS). He is also an Editorial Board Member for Information Fusion and the ISPRS Journal of Photogrammetry and Remote Sensing. He has received several prestigious awards, including the Jose Bioucas Dias Award (2021) and Paul Gader Award (2024) at WHISPERS for outstanding papers, respectively, the Remote Sensing Young Investigator Award (2022), the IEEE GRSS Early Career Award (2022), and the ``2023 China’s Intelligent Computing Innovators'' award (the only recipient in AI for Earth Science) by MIT Technology Review (2024). He has been recognized as a Highly Cited Researcher by Clarivate Analytics in 2022, 2023, and 2024.
\end{IEEEbiography}

\begin{IEEEbiography}[{\includegraphics[width=1in,height=1.25in,clip,keepaspectratio]{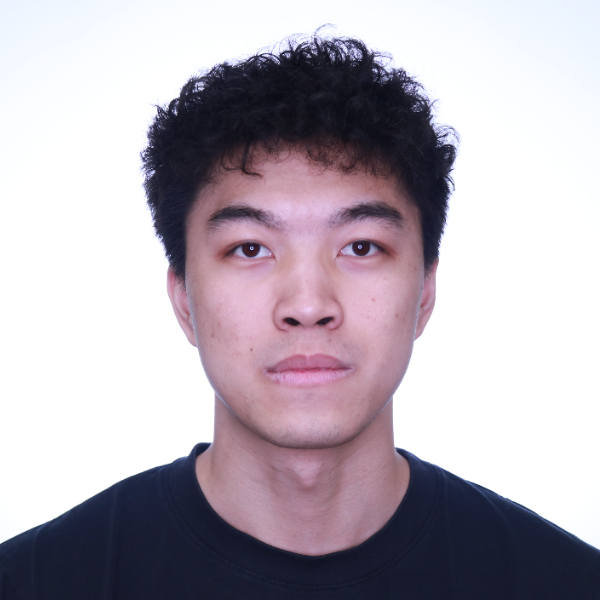}}]{Shuhang Ge} received his B.S. degree in Century College, Beijing University of Posts and Telecommunications in 2022, and his master degree from New York University in 2024. He is currently a Machine Learning Engineer at ByteDance. His research interests include Large Language Models, Multimodal Artificial Intelligence and Generation Models.
\end{IEEEbiography}

\begin{IEEEbiography}[{\includegraphics[width=1in,height=1.25in,clip,keepaspectratio]{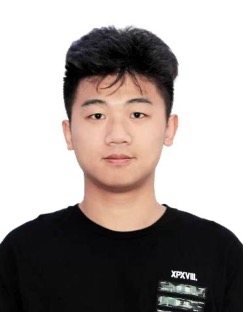}}]{Chuyao Luo} is currently a Associate Researcher at Harbin Institute of Technology (Shenzhen). He received his Bachelor's degree in Internet of things Engineering from Dalian Maritime University. He received his Doctor's degree in Computer Science at Harbin Institute of Technology (Shenzhen). His research includes data mining, computer vision, time-series data prediction, and precipitation nowcasting.
\end{IEEEbiography}

\begin{IEEEbiography}[{\includegraphics[width=1in,height=1.25in,clip,keepaspectratio]{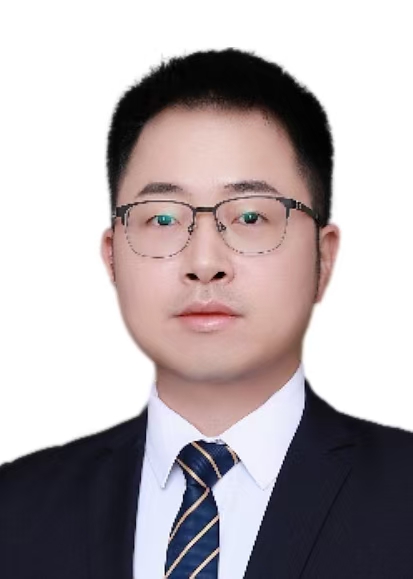}}]{Kai Jiang} received Ph.D. degree from the University of Science and Technology of China, Anhui, China, in 2014. Kai Jiang is a Senior Engineer in Nanjing Research Institute of Electronic Engineering. He is the head of Key Lab of Information System Requirement. His main research interests are systems engineering, data mining, and large scale system architecture.

\end{IEEEbiography}

\begin{IEEEbiography}[{\includegraphics[width=1in,height=1.25in,clip,keepaspectratio]{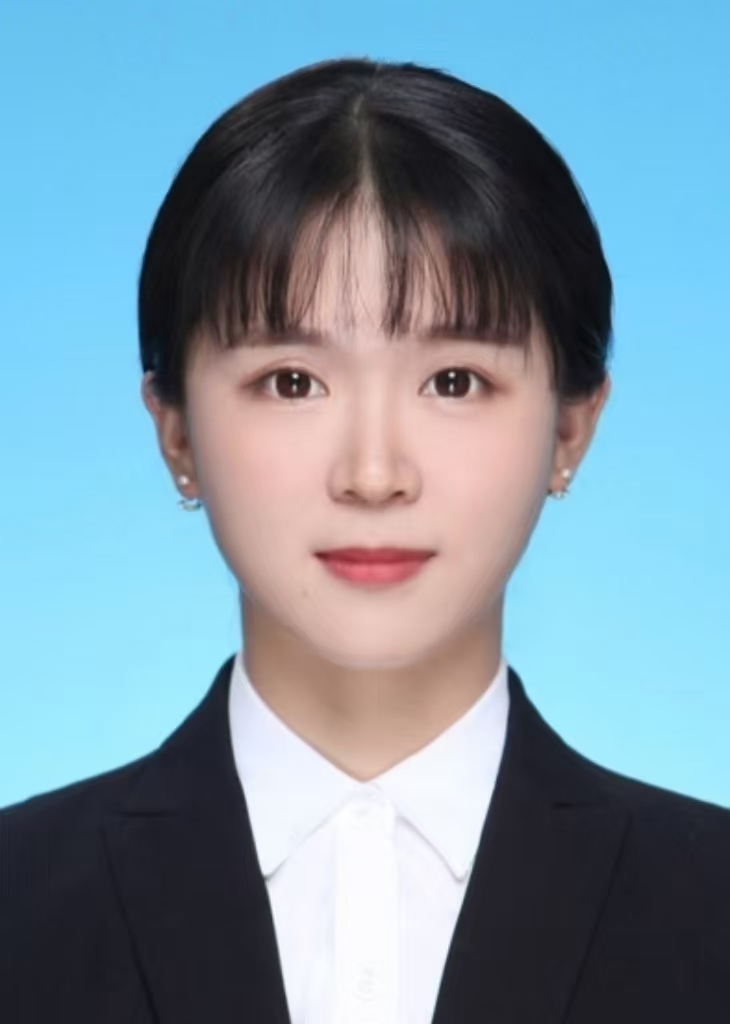}}]{Hao Jin} received Ph.D. degree from Institute of Information Engineering, Chinese Academy of Sciences, Beijing, China, in 2018. Hao Jin is a Senior Engineer in National Engineering Research Center for Public Safety Risk Perception and Control by Big Data (RPP), China Academic of Electronics and Information Technology. Her main research interests are artificial intelligence and big data.

\end{IEEEbiography}

\begin{IEEEbiography}[{\includegraphics[width=1in,height=1.25in,clip,keepaspectratio]{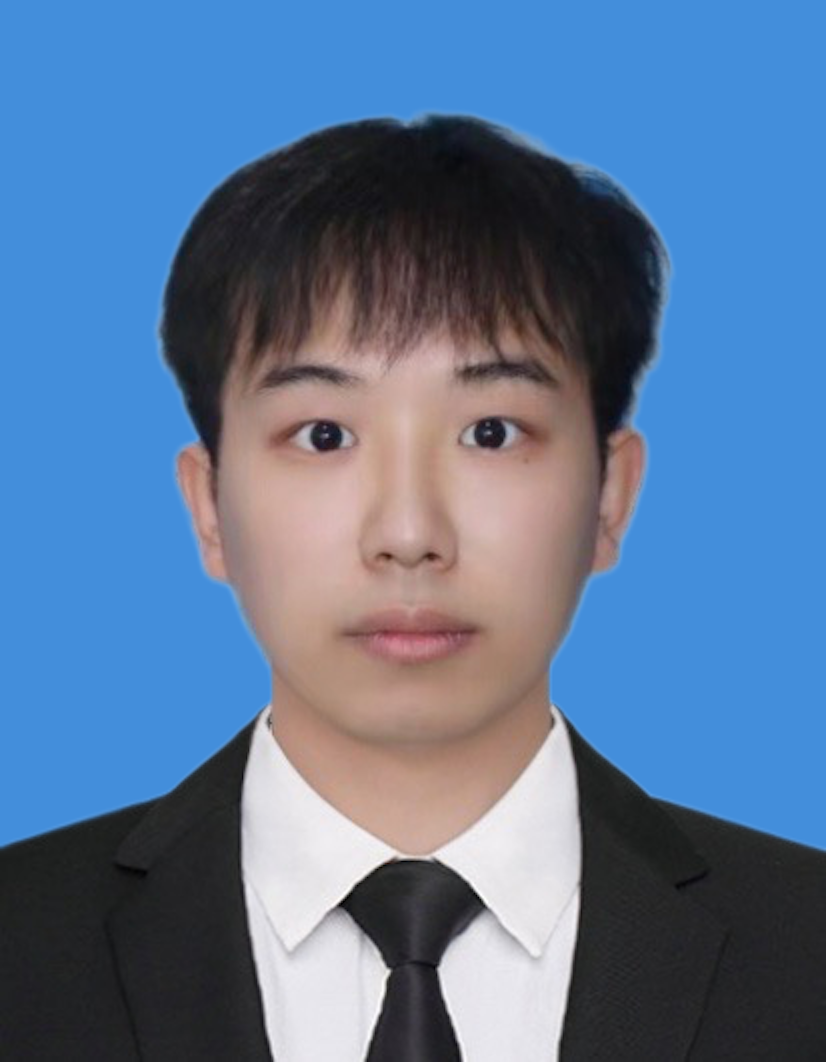}}]{Congcong Wen} (Member, IEEE) received his B.S. degree in Geographic Information Systems from the China University of Petroleum, China, in 2016, and his Ph.D. degree from the Aerospace Information Research Institute, Chinese Academy of Sciences, China, in 2021. He is currently a Postdoctoral Associate with the Department of Electrical and Computer Engineering at New York University and New York University Abu Dhabi. His research interests include multimodal artificial intelligence, 3D computer vision, foundation models, and remote sensing.
\end{IEEEbiography}

\end{document}